\documentclass{article}

\usepackage{microtype}
\usepackage{graphicx}
\usepackage{subcaption}
\usepackage{booktabs} 

\usepackage{hyperref}

\usepackage[accepted]{icml2026}

\usepackage{amsmath}
\usepackage{amssymb}
\usepackage{mathtools}
\usepackage{amsthm}

\usepackage[capitalize,noabbrev]{cleveref}

\usepackage{xspace}
\usepackage{multirow}
\usepackage{paralist}
\usepackage{lipsum}
\usepackage{arydshln}
\usepackage[skins]{tcolorbox}
\usepackage{xcolor}
\usepackage{wrapfig}

\theoremstyle{plain}
\newtheorem{theorem}{Theorem}[section]

\theoremstyle{definition}
\newtheorem{definition}[theorem]{Definition}

\theoremstyle{remark}

\definecolor{usercolor}{RGB}{0, 111, 120} 
\definecolor{modelcolor}{RGB}{99, 80, 210}
\definecolor{examplecolor}{RGB}{220, 77, 0}
\definecolor{backcolor}{RGB}{249, 250, 251}
\definecolor{framecolor}{RGB}{2, 40, 78}

\usepackage[textsize=tiny]{todonotes}

\icmltitlerunning{Generative Representation Learning on Hyper-relational Knowledge Graphs via Masked Discrete Diffusion}

\renewcommand{\vec}[1]{{\bf{#1}}}

\newcommand{\RR}{\mathbb{R}}
\newcommand{\EE}{\mathbb{E}}

\providecommand{\vc}{\ensuremath{\vec{c}}}
\providecommand{\ve}{\ensuremath{\vec{e}}}
\providecommand{\vm}{\ensuremath{\vec{m}}}
\providecommand{\vp}{\ensuremath{\vec{p}}}
\providecommand{\vx}{\ensuremath{\vec{x}}}
\providecommand{\vz}{\ensuremath{\vec{z}}}

\newcommand{\set}[1]{\mathcal{#1}}
\providecommand{\sC}{\ensuremath{\set{C}}}
\providecommand{\sD}{\ensuremath{\set{D}}}
\providecommand{\sH}{\ensuremath{\set{H}}}
\providecommand{\sL}{\ensuremath{\set{L}}}
\providecommand{\sM}{\ensuremath{\set{M}}}
\providecommand{\sO}{\ensuremath{\set{O}}}
\providecommand{\sQ}{\ensuremath{\set{Q}}}
\providecommand{\sR}{\ensuremath{\set{R}}}
\providecommand{\sU}{\ensuremath{\set{U}}}
\providecommand{\sV}{\ensuremath{\set{V}}}

\newcommand{\mrr}[0]{\textup{MRR}\xspace}
\newcommand{\hten}[0]{\textup{Hit10}\space}
\newcommand{\hone}[0]{\textup{Hit1}\space}

\newcommand{\wdk}[0]{\textup{WD50K}\xspace}
\newcommand{\wppm}[0]{\textup{WikiPeople$^-$}\xspace}
\newcommand{\wpp}[0]{\textup{WikiPeople}\xspace}

\newcommand{\ours}[0]{\textup{KREPE}\xspace}

\newcommand{\kicgpt}[0]{\textup{KICGPT}\xspace}
\newcommand{\mukdc}[0]{\textup{MuKDC}\xspace}

\newcommand{\maypl}[0]{\textup{MAYPL}\xspace}

\newcommand{\hyperf}[0]{\textup{HyperFormer}\xspace}
\newcommand{\hahe}[0]{\textup{HAHE}\xspace}
\newcommand{\stare}[0]{\textup{StarE}\xspace}
\newcommand{\hinge}[0]{\textup{HINGE}\xspace}
\newcommand{\gran}[0]{\textup{GRAN}\xspace}
\newcommand{\hytrans}[0]{\textup{Hy-Transformer}\xspace}
\newcommand{\hynt}[0]{\textup{HyNT}\xspace}
\newcommand{\shrinke}[0]{\textup{ShrinkE}\xspace}
\newcommand{\nalp}[0]{\textup{NaLP}\xspace}
\newcommand{\hdiff}[0]{\textup{HDiff}\xspace}
\newcommand{\fognn}[0]{\textup{FormerGNN}\xspace}

\newcommand{\rae}[0]{\textup{RAE}\xspace}
\newcommand{\hype}[0]{\textup{HypE}\xspace}
\newcommand{\hyconve}[0]{\textup{HyConvE}\xspace}
\newcommand{\mseakg}[0]{\textup{MSeaKG}\xspace}
\newcommand{\hje}[0]{\textup{HJE}\xspace}
\newcommand{\hycube}[0]{\textup{HyCubE}\xspace}
\newcommand{\ram}[0]{\textup{RAM}\xspace}
\newcommand{\sts}[0]{\textup{S2S}\xspace}
\newcommand{\tnalp}[0]{\textup{tNaLP}\xspace}
\newcommand{\neuinf}[0]{\textup{NeuInfer}\xspace}
\newcommand{\polye}[0]{\textup{PolygonE}\xspace}
\newcommand{\hcnet}[0]{\textup{HCNet}\xspace}
\newcommand{\hysae}[0]{\textup{HySAE}\xspace}
\newcommand{\gahe}[0]{\textup{GAHE}\xspace}

\begin{document}

\twocolumn[
  \icmltitle{Generative Representation Learning on Hyper-relational Knowledge Graphs via Masked Discrete Diffusion}




  



  \icmlsetsymbol{equal}{*}

  \begin{icmlauthorlist}
    \icmlauthor{Jaejun Lee}{kaistcs}
    \icmlauthor{Seheon Kim}{kaistcs}
    \icmlauthor{Joyce Jiyoung Whang}{kaistaic}
  \end{icmlauthorlist}

  \icmlaffiliation{kaistcs}{School of Computing,}
  \icmlaffiliation{kaistaic}{Department of AI Computing, KAIST, Daejeon, South Korea}

  \icmlcorrespondingauthor{Joyce Jiyoung Whang}{jjwhang@kaist.ac.kr}

  \icmlkeywords{Hyper-relational Knowledge Graph, Fact Generation, Generative Representation Learning, Knowledge Graph Completion, Link Prediction, Message Passing Neural Network}

  \vskip 0.3in
]



\printAffiliationsAndNotice{}  


\begin{abstract}
   Hyper-relational knowledge graphs (HKGs) effectively represent complex facts. While inferring new knowledge in HKGs is a critical problem, current methods cast it as a simple link prediction, assuming that nearly all entities and relations within a fact are known, leaving only a single blank to be filled. However, this restricted assumption may not hold in real-world scenarios in which multiple, or even all, constituent components of a fact may be missing simultaneously. To bridge this gap, we introduce a task called fact generation: generating a valid hyper-relational fact from an arbitrarily masked query, i.e., completing a partially observed fact or generating a fact from scratch. We propose \ours, the first generative representation learning method for HKGs that learns to model the probability distributions of missing components conditioned on the local fact components and global structure of HKGs via a masked discrete diffusion. \ours models both the intra-fact dependencies by contextual message passing and inter-fact correlations by aggregating stochastically sampled contexts. \ours seamlessly unifies link prediction and fact generation within a single training framework, achieving state-of-the-art performance on standard HKG link prediction benchmarks and outperforming LLM-based baselines in generating novel and correct facts.
\end{abstract}

\section{Introduction}

\begin{figure}[t]
\centering
\includegraphics[width=0.95\linewidth]{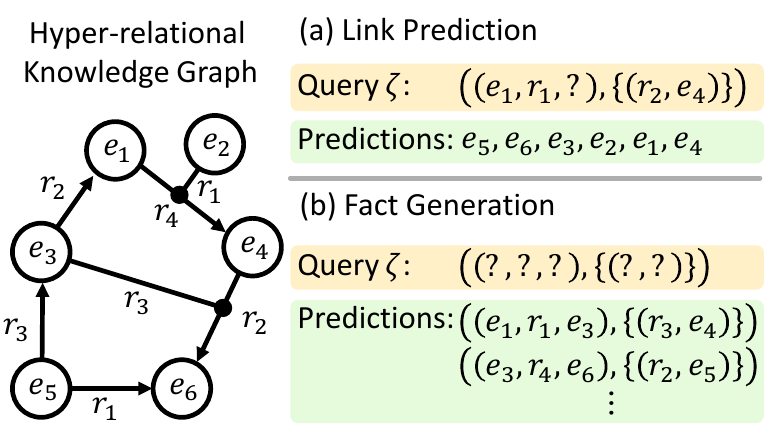}
\captionof{figure}{Comparison of link prediction and fact generation tasks on HKGs. Link prediction retrieves a single missing element by ranking candidates, whereas fact generation constructs valid facts from arbitrarily masked inputs, including fully masked ones.}
\label{fig:cmp}
\end{figure}

Knowledge graphs (KGs) structure human knowledge by representing facts as triplets. However, the simple triplet format is insufficient to capture the multifaceted nature of real-world facts. To address this limitation, knowledge bases such as Wikidata~\cite{wk} and YAGO~\cite{yago} extend the triplets into hyper-relational facts by adding auxiliary key-value pairs, known as qualifiers, forming hyper-relational knowledge graphs (HKGs). Since real-world HKGs are often incomplete, link prediction has been proposed as an HKG-completion task, where each problem is defined as predicting a single missing element in a fact, given all the other elements. As illustrated in Figure~\ref{fig:cmp}(a), a link prediction method outputs a ranked list of candidates for the target missing element, assuming that all the other entities and relations within the fact are known.

Some link prediction methods assess the plausibility of each fact formed by filling the missing entry with a candidate component~\cite{hinge, neuinf}, and other methods employ the transformer architecture to predict a missing component in an incomplete fact~\cite{hynt, gran}. However, these methods focus on processing facts individually and do not sufficiently consider the complex dependencies between facts within the HKG structure. While approaches combining GNN-based encoding with Transformer decoding have been proposed~\cite{stare, hahe}, they use the graph structure in a limited capacity, for instance, by ignoring relations or qualifiers in updating representations. Recently, \maypl~\cite{maypl} has been introduced to comprehensively encode the HKG structure into representation learning.

Despite these architectural advancements, a fundamental limitation persists across existing works: they are restricted to the task of link prediction, which is essentially a passive ``fill-in-the-blank" problem relying on the rigid assumption of a single missing element. This assumption is inconsistent with real-world scenarios, where the extent of missing information in queries varies unpredictably. In practice, we are rarely presented with a nearly complete fact requiring a single component to be filled, e.g., $\big((h,r,\texttt{?}), \{(k_1,v_1)\}\big)$. Instead, the challenge often lies in generating valid knowledge from highly incomplete observations. To accommodate the various levels of missing information in real-world queries, we propose a task named \textit{Fact Generation}. Unlike link prediction, which infers a specific missing component based on an almost-complete query with a single mask, fact generation aims to construct a valid hyper-relational fact from an arbitrarily masked query. As shown in Figure~\ref{fig:cmp}(b), our task extends even to generating complete facts from a fully masked input $\big((\texttt{?},\texttt{?},\texttt{?}),\{(\texttt{?},\texttt{?})\}\big)$.

In this work, we formulate fact generation as the task of modeling the conditional probability distribution of masked components given a query and the HKG structure. We present \ours (Contextual H\textbf{K}G \textbf{REP}resentation Learning via Masked Discret\textbf{E} Diffusion), which leverages contextual message passing to compute entity and relation representations and employs masked discrete diffusion to estimate the underlying distribution of masked components. Specifically, our contextual message passing updates each component to encode its surrounding structure, i.e., its context within each fact, capturing intra-fact dependencies. Simultaneously, a bi-level noising strategy is employed to simulate diverse conditional generation scenarios for both the query and the observed structure. \ours is trained under a masked discrete diffusion framework that treats generation as an iterative reconstruction process, thereby learning the joint conditional distribution of multiple masked components within a query. Since link prediction is a special case of fact generation involving a single masked component, \ours seamlessly unifies entity prediction, relation prediction, and fact generation within a single representation learning framework. Our main contributions are summarized as follows:
\begin{compactitem}
\item We propose \textit{Fact Generation}, expanding the scope of HKG completion from passive ranking of candidates for single blanks to direct generation of missing facts.
\item We present \ours, the first generative representation learning method for HKGs incorporating masked discrete diffusion to explicitly model the joint conditional probability distributions of missing components in a query. Our code and datasets are available at \url{https://github.com/bdi-lab/KREPE}.
\item \ours achieves state-of-the-art performance on three standard link prediction benchmarks, demonstrating that our generative objective enhances the quality of learned representations.
\item \ours significantly outperforms a diverse set of competitive LLM-based baselines in fact generation, validating that modeling the probability distributions conditioned on both inter-fact and intra-fact dependencies is effective for generating novel and correct facts.
\end{compactitem}

\section{Related Work}
\paragraph{HKG Representation Learning} Most HKG representation learning methods~\cite{gran, maypl} are tailored for link prediction. Scoring-based methods such as \hinge~\cite{hinge} and \neuinf~\cite{neuinf} rank all possible candidates for a single missing position. These approaches are computationally intractable for fact generation involving multiple missing components, as the combinatorial search space grows exponentially. \stare~\cite{stare} and \hytrans~\cite{hytrans} encode an incomplete fact into a vector to predict a single target, and \hyperf~\cite{hyperf} requires the neighbor information of observed entities, restricting them to single-element inference. Although \hahe~\cite{hahe} introduces a multi-position prediction task, it is constrained by predefined mask patterns, allowing at most one missing component per triplet or qualifier pair. As such, existing methods are limited to predicting single missing elements or to handling fixed, limited patterns of masks and lack the capability to support knowledge generation in real-world scenarios with varying mask patterns.
\paragraph{LLMs for KGs.}
Although focused on KGs rather than HKGs, efforts have been made to adapt large language models (LLMs) for KG completion, particularly for link prediction~\cite{mpikgc, mkgl, kgfit}. A common approach to adapting LLMs to KG link prediction is a \textit{retrieve-then-rerank} strategy~\cite{dift, sat, ssqr}, where a pre-trained KG model retrieves candidates for a single-blank query, which are subsequently refined by an LLM. For instance, \kicgpt~\cite{kicgpt} utilizes RotatE~\cite{rotate} to get top-$k$ candidates for a given query and employs an LLM to re-rank them. However, these methods require a backbone KG link prediction model to provide candidates, and thus cannot be directly applied to fact generation, as existing models for KGs cannot handle multi-masked queries. While \mukdc~\cite{mukdc} synthesizes triplets using LLMs, it is a data augmentation technique to mitigate sparsity in few-shot settings, rather than a generative model for inferring novel facts during inference. Obviously, we also distinguish our work from text-to-KG construction methods~\cite{kggen, textnkg}; while they aim to convert unstructured text corpora into structured knowledge, our objective is to generate new knowledge, which requires inference capability on HKGs fundamentally distinct from simple information extraction.
\paragraph{Diffusion Models for (H)KGs}
Denoising diffusion probabilistic models~\cite{ddpm} have become a dominant regime for generative modeling on continuous data, such as images. To extend them to discrete data, discrete diffusion methods~\cite{dddpm}, such as masked discrete diffusion, have been developed for tasks like text generation. While several diffusion-based methods have been proposed in the KG domain~\cite{kgdm, fdm}, they are conditional denoising diffusion models that generate a continuous embedding vector for a missing entity conditioned on a query, rather than generating the fact itself. DiffusionE~\cite{diffe} utilizes heat diffusion on Riemannian manifolds to reformulate message passing for link prediction, which fundamentally differs from the generative denoising diffusion. \hdiff~\cite{hdiff} applies denoising diffusion to HKGs to refine continuous fact representations for link prediction. While existing methods use diffusion merely to refine continuous embeddings for candidate ranking, \ours leverages the diffusion process to model the joint probability distribution of missing components in a query, enabling the generation of facts from arbitrarily masked queries.
\paragraph{Triple Set Prediction}Triple set prediction~\cite{tsp} is the task of predicting missing triplets in which all elements are unknown. However, this task is defined for KGs and does not account for scenarios where partial information about missing triplets is available. In contrast, fact generation is defined for HKGs and aims to predict valid facts from arbitrarily masked queries, including fully masked ones. Thus, triple set prediction is a special case of fact generation. Notably, GPHT~\cite{tsp}, a model for triple set prediction, cannot be directly applied to fact generation. GPHT reduces the number of candidate triplets and then ranks the remaining candidate triplets for prediction. However, for hyper-relational facts, the candidate space grows exponentially with the number of qualifiers, making such ranking-based prediction impractical.


\begin{figure*}[t]
\centering
\includegraphics[width=0.98\linewidth]{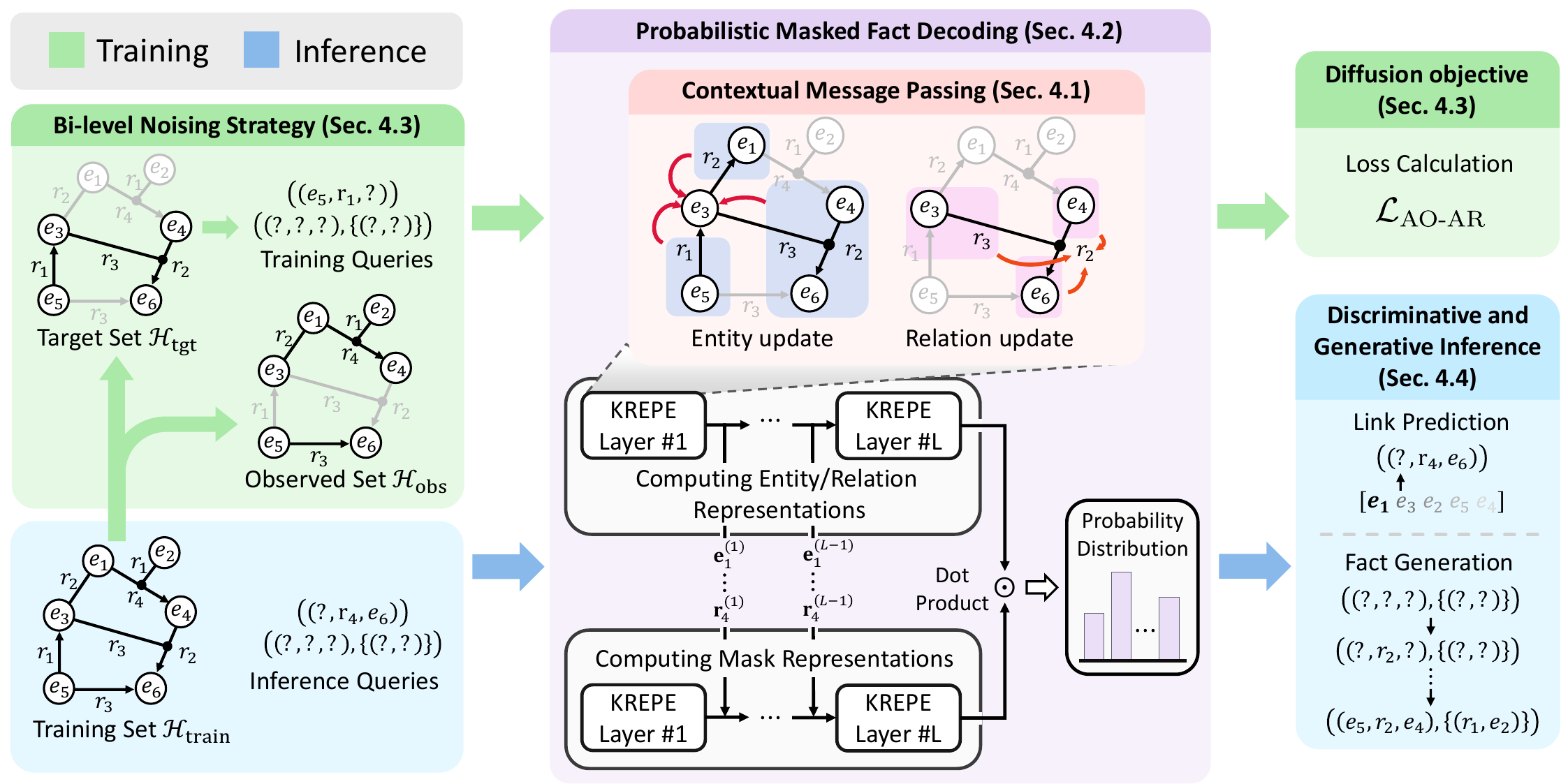}
\captionof{figure}{Overview of \ours. During training, the training set is partitioned into an observed set and a target set, where the facts in the target set are masked to serve as queries. Representations of the entities, relations, and masks are updated via contextual message passing, integrating both the local contexts within the facts and the relationship between facts. These representations are used to approximate the probability distribution of the masks. The estimated probability distribution is used for the diffusion objective. During inference, the training set acts as the observed set, and the resulting probability distribution is used for both discriminative and generative inference tasks.}

\label{fig:overview}
\end{figure*}
\section{Problem Definition}
\label{sec:def}
We formally define HKGs (Definition~\ref{def:hkg}) and introduce the tasks of link prediction (Definition~\ref{def:lp}) and fact generation (Definition~\ref{def:fg}) on HKGs.

\begin{definition}[Hyper-relational Knowledge Graph]
\label{def:hkg}
A hyper-relational knowledge graph (HKG) is defined as $G=(\sV, \sR, \sH)$, where $\sV$ and $\sR$ denote the sets of entities and relations, respectively, and $\sH$ denotes the set of hyper-relational facts. A hyper-relational fact $\xi\in\sH$ comprises a primary triplet and a set of qualifiers, denoted as $\xi=\big((h,r,t), \sQ_\xi\big)$. The primary triplet $(h,r,t)$ consists of a head entity $h\in\sV$, a primary relation $r\in\sR$, and a tail entity $t\in\sV$. The set of qualifiers is defined as $\sQ_\xi=\{(k_{i},v_{i})\mid k_i\in\sR,v_i\in\sV\}_{i=1}^{n_\xi}$, where $n_\xi\geq0$ is the number of qualifiers in $\xi$, and each qualifier pair ($k_i$, $v_i$) consists of a qualifier relation $k_i$ and a qualifier entity $v_i$.
\end{definition}

\begin{definition}[Link Prediction on HKGs]
\label{def:lp}
Given an HKG $G=(\sV,\sR,\sH)$, let $\xi$ be a valid hyper-relational fact composed of entities from $\sV$ and relations from $\sR$. Let $\zeta$ be an incomplete hyper-relational fact derived from $\xi$ by masking a single component $y$. Link prediction is the task of inferring the missing component $y$. Link prediction is categorized by the type of $y$: if $y\in\sV$, the task is \textit{entity prediction}, and if $y\in\sR$, the task is \textit{relation prediction}.
\end{definition}

\begin{definition}[Fact Generation on HKGs]
\label{def:fg}
Given an HKG $G=(\sV,\sR,\sH)$, let $\xi$ be a valid hyper-relational fact composed of entities and relations from $\sV$ and $\sR$, respectively. Let $\zeta$ be an incomplete hyper-relational fact derived from $\xi$ by masking an arbitrary subset of its components. Fact generation is the task of generating a valid fact $\xi'$ from $\zeta$ by replacing the masked components with entities in $\sV$ and relations in $\sR$. Note that $\xi'$ is not necessarily identical to $\xi$.
\end{definition}

\section{\ours: Contextual HKG Representation Learning via Masked Discrete Diffusion}
\label{sec:ours}

We propose \ours, which utilizes contextual message passing to compute representations and is trained via masked discrete diffusion to gain generative capabilities. To update entity and relation representations, \ours aggregates messages from the facts they belong to while explicitly excluding their own information from those facts' messages; this exclusion forces their representations to rely solely on the surrounding ``context" in the facts. The representation of a masked component is also derived from its context and compared against all candidate representations to compute its probability distribution. Figure~\ref{fig:overview} illustrates the training and inference pipeline. During training, \ours partitions the training set into an observed set and a target set; facts in the observed set are used to compute entity and relation representations, while the facts in the target set are masked and serve as queries. In contrast, the entire training set serves as the observed set during inference. The probability distributions of the masked components are used to compute the diffusion loss during training and to perform both discriminative and generative tasks during inference. Implementation details of \ours are provided in Appendix~\ref{app:imp}.

\subsection{Contextual Message Passing}
\label{subsec:layer}
To compute representations for entities and relations, \ours employs contextual message passing, in which the message from a fact to each of its constituent components is computed using the fact's components, excluding the target component itself. The remaining components that do not include the target component within a fact are termed the ``context'' of the target component. By removing the component's own representation from the messages, \ours prevents the trivial retention of self information while enforcing reliance on the other components. 

 Let $\vp^{(l)},\ve^{(l)}\in\RR^d$ denote the representation of a relation $p$ and an entity $e$ at the $l$-th layer, respectively, where $d$ denotes the dimension, $l\in\{1,\dots,L\}$, and $L$ is the number of layers. The representation of an entity $e$ is initialized by $\ve^{(0)}=\vz_\mathrm{ENT}$, where $\vz_\mathrm{ENT}\in\RR^d$ is a shared learnable token for all entities. Similarly, the representation of a relation $p$ is initialized by $\vp^{(0)}=\vz_\mathrm{REL}$, where $\vz_\mathrm{REL}\in\RR^d$ is a shared learnable token for all relations.

Let us decompose a fact into a set of its constituent relation-entity pairs. For example, a hyper-relational fact $\xi=\big((h,r,t),\sQ_\xi\big)$ is decomposed into the set of pairs $\{(r,h), (r,t), (k_1, v_1), \cdots,(k_{n_\xi},v_{n_\xi})\}$, where $(k_i,v_i)\in\sQ_\xi$. Then, each pair is associated with a specific role: head, tail, or qualifier, based on the position of the entity, i.e., head entity, tail entity, or qualifier entity. For example, the role of $(r,h)$ is head, $(r,t)$ is tail, and $(k_i,v_i)$ is qualifier. Let $(p,e)$ denote a generic relation-entity pair associated with a role $\rho\in\{\texttt{head}, \texttt{tail}, \texttt{qual}\}$ in $\xi$, where $p$ is a relation and $e$ is an entity. The representation $\vz_{(p,e)}^{(l)}$ of the pair $(p,e)$ at the $l$-th layer is defined as follows:
\begin{equation}
\vz_{(p,e)}^{(l)}=W^{(l)}_\rho[\vp^{(l-1)};\ve^{(l-1)}]
\end{equation}
where $W^{(l)}_\texttt{head}, W^{(l)}_\texttt{tail}, W^{(l)}_\texttt{qual}\in\RR^{d\times 2d}$ are the projection matrices for the respective roles and $[;]$ denotes concatenation. The representation of a fact $\xi$ at layer $l$, denoted as $\vz_\xi^{(l)}$, is obtained by summing its constituent pairs:
\begin{equation}
\vz_\xi^{(l)}=\vz_{(r,h)}^{(l)} + \vz_{(r,t)}^{(l)} + \sum_{i=1}^{n_\xi}\vz_{(k_i,v_i)}^{(l)} \label{eq:fact}
\end{equation}
To update a component within a fact, we compute a \textit{context message} that encodes all the other components in the fact. This ensures that the update is conditioned on the surrounding structure of the component. The context messages $\vm_{\xi\to e}^{(l)}$ and $\vm_{\xi\to p}^{(l)}$ from fact $\xi$ to its components $e$ and $p$, respectively, in the constituent pair $(p,e)$ are computed by:
\begin{gather}
\vm_{\xi\to (p,e)}^{(l)} = \text{MLP}^{(l)}\left(\frac{\vz_\xi^{(l)}-\vz_{(p,e)}^{(l)}}{n_\xi+1}\right)\\
\vm_{\xi\to e}^{(l)} = W^{(l)}_\mathrm{ENT}[\vm_{\xi\to(p,e)}^{(l)};\vp^{(l-1)}] \label{eq:cme}\\
\vm_{\xi\to p}^{(l)} = W^{(l)}_\mathrm{REL}[\vm_{\xi\to(p,e)}^{(l)};\ve^{(l-1)}] \label{eq:cmr}
\end{gather}
where $\text{MLP}$ denotes a multi-layer perceptron and $W^{(l)}_\mathrm{ENT}$, $W^{(l)}_\mathrm{REL}\in\RR^{d\times2d}$ denote the projection matrices. The message $\vm_{\xi\to (p,e)}^{(l)}$ from $\xi$ to $(p,e)$ represents the composition of all the other pairs within $\xi$. The context message $\vm_{\xi\to e}^{(l)}$ is formed by projecting the concatenation of this message with the representation of the paired relation $p$. This allows the context of $e$ within $\xi$ to incorporate all complementary components of $e$. Similarly, $\vm_{\xi\to p}^{(l)}$ is computed by projecting the concatenation of $\vm_{\xi\to (p,e)}^{(l)}$ with $\ve^{(l-1)}$.

Finally, the representation of an entity $e$ is updated from $\ve^{(l-1)}$ to $\ve^{(l)}$ by aggregating the context messages from all facts involving $e$ using a multi-head attention mechanism followed by an MLP~\cite{attn}. A similar process is applied to update the relation representations using distinct parameters. By iteratively aggregating the context messages across layers, the final representations encode the structural dependencies between the facts.

\subsection{Probabilistic Masked Fact Decoding}
\label{subsec:pd}
\ours is designed to generate facts from an incomplete fact with an arbitrary number of unknown components. Formally, an incomplete fact $\zeta$ is defined as a fact where one or more, potentially all, of its components are unknown, i.e., masked. Each masked entity and relation is initialized with a learnable vector, $\vx_\text{ENT}\!\in\!\RR^d$ and $\vx_\text{REL}\!\in\!\RR^d$, respectively.

Given an HKG $G=(\sV,\sR,\sH)$ and a masked fact $\zeta$, \ours treats each masked entity and relation as a vanilla entity and relation, respectively, and updates their representations via contextual message passing. In this process, $\zeta$ is not used to update the known components; $\zeta$ is used only to update its masked components. Specifically, the context message of a masked component $x$ in $\zeta$ at the $l$-th layer is computed using the $(l\!-\!1)$-th layer's representations of known entities and relations in $\zeta$ and the $(l\!-\!1)$-th layer's representations of the other masked components in $\zeta$, if any. This allows the representation of $x$ to be conditioned on the context from $\zeta$. Let $\vx_\text{mask}^{(L)}$ denote the final representation of $x$. \ours models the probability distribution of $x$ on the candidate set $\sC$, either $\sV$ or $\sR$ depending on $x$, as:
\begin{equation}
\small
\label{eq:pd}
P_\theta(x\mid \zeta, G) = \underset{c\in\sC}{\text{Softmax}} \left(\vx_\text{mask}^{(L)} \cdot \vc^{(L)}\right)
\end{equation}
where $\vc^{(L)}$ is the final representation of a candidate $c$ computed using the facts in $G$. Our decoding strategy enables \ours to estimate the probability distribution of missing components by integrating two complementary sources of information: (1) the local context of the masked component in the query, embedded in $\vx_\text{mask}^{(L)}$, and (2) the global HKG structure, encoded across the candidate representations $\vc^{(L)}$.

\begin{table*}[t]
\parbox{0.6\linewidth}{
\centering
\footnotesize
\caption{Entity prediction results on \wdk and \wppm. Pri and All denote prediction on primary triplets and all positions including qualifiers, respectively.}
\label{tb:hkgent}
\setlength{\tabcolsep}{0.5em}
\begin{tabular}{ccccccccc}
\toprule
& & \multicolumn{3}{c}{\wdk} & \multicolumn{3}{c}{\wppm} \\
 \cmidrule(lr){3-5} \cmidrule(lr){6-8}
&  & \mrr & \hten & \hone & \mrr & \hten & \hone \\
\midrule
\multirow{12}{*}{Pri} & \stare~(\citeyear{stare}) & 0.349 & 0.496 & 0.271 & 0.491 & 0.648 & 0.398 \\
& \gran~(\citeyear{gran}) & -      & -      & - & 0.503 & 0.620 & 0.438 \\
& \hytrans~(\citeyear{hytrans}) & 0.356 & 0.498 & 0.281 & 0.501 & 0.634 & 0.426 \\
& \hyperf~(\citeyear{hyperf}) & 0.366 & 0.514 & 0.288 & 0.473 & 0.646 & 0.361 \\
& \hahe~(\citeyear{hahe}) & 0.368 & 0.516 & 0.291 & 0.509 & 0.639 & \textbf{0.447} \\
& \shrinke~(\citeyear{shrinke}) & 0.345 & 0.482 & 0.275 & 0.485 & 0.601 & 0.431 \\
& \hynt~(\citeyear{hynt}) & 0.357 & 0.501 & 0.281 & - & - & - \\
& \hdiff~(\citeyear{hdiff}) & 0.373 & 0.512 & 0.292 & \underline{0.512} & 0.642 & \textbf{0.447} \\
& \fognn~(\citeyear{fognn}) & 0.377 & 0.527 & \underline{0.299} & 0.494 & 0.653 & 0.392 \\
& \maypl~(\citeyear{maypl}) & \underline{0.381} & \underline{0.544} & 0.297 & \textbf{0.519} & \underline{0.657} & \underline{0.444} \\
\cmidrule(lr){2-8}
& \ours & \textbf{0.389} & \textbf{0.552} & \textbf{0.303} & \textbf{0.519} & \textbf{0.663} & 0.437 \\
\midrule
\multirow{5}{*}{All} & \hahe~(\citeyear{hahe}) & 0.402 & 0.546 & \underline{0.327} & 0.495 & 0.631 & 0.420 \\
& \hynt~(\citeyear{hynt}) & 0.383 & 0.527 & 0.308 & - & - & - \\
& \hdiff~(\citeyear{hdiff}) & 0.405 & 0.546 & 0.325 & 0.502 & 0.639 & 0.430 \\
& \maypl~(\citeyear{maypl}) & \underline{0.411} & \underline{0.572} & 0.326 & \underline{0.521} & \underline{0.659} & \textbf{0.446} \\
\cmidrule(lr){2-8}
& \ours & \textbf{0.419} & \textbf{0.580} & \textbf{0.334} & \textbf{0.522} & \textbf{0.666} & \underline{0.439} \\
\bottomrule
\end{tabular}
}
\hfill
\parbox{0.36\linewidth}{
\centering
\footnotesize
\caption{Entity prediction results on all entities within facts in \wpp.}
\label{tb:wppent}
\begin{tabular}{ccccc}
\toprule
 & \multicolumn{3}{c}{\wpp} \\
 \cmidrule(lr){2-4}
 & \mrr & \hten & \hone \\
\midrule
\rae~(\citeyear{rae}) & 0.172 & 0.320 & 0.102 \\ 
\nalp~(\citeyear{nalp}) & 0.338 & 0.466 & 0.272 \\
\hype~(\citeyear{hype}) & 0.282 & 0.487 & 0.148 \\ 
\neuinf~(\citeyear{neuinf}) & 0.350 & 0.467 & 0.282 \\
\ram~(\citeyear{ram}) & 0.380 & 0.539 & 0.279 \\
\sts~(\citeyear{sts}) & 0.372 & 0.533 & 0.277 \\
\polye~(\citeyear{polye}) & 0.431 & 0.568 & 0.334 \\
\hyconve~(\citeyear{hyconve}) & 0.362 & 0.501 & 0.275 \\
\mseakg~(\citeyear{mseakg}) & 0.392 & 0.553 & 0.290 \\
\tnalp~(\citeyear{tnalp})  & 0.350 & 0.471 & 0.288 \\
\hcnet~(\citeyear{hcnet}) & 0.421 & 0.565 & 0.344 \\ 
\hycube~(\citeyear{hycube}) & 0.448 & 0.592 & 0.368  \\
\hje~(\citeyear{hje}) & 0.450 & 0.582 & 0.375 \\
\gahe~(\citeyear{gahe}) & 0.415 & 0.562 & 0.320 \\
\hysae~(\citeyear{hysae}) & 0.454 & 0.603 & 0.373 \\
\maypl~(\citeyear{maypl}) & \underline{0.488} & \underline{0.635} &  \underline{0.405}\\
\midrule
\ours & \textbf{0.491} & \textbf{0.642} & \textbf{0.406} \\
\bottomrule
\end{tabular}
}
\end{table*}

\subsection{Generative Training via Masked Discrete Diffusion}
\label{subsec:tr}
To model the distribution of a masked component $x$ in a query $\zeta$ conditioned on both the query $\zeta$ and the HKG $G$, i.e., $P_\theta(x|\zeta, G)$, \ours utilizes masked discrete diffusion~\cite{dddpm}. However, unlike standard discrete diffusion where the corrupted input is often the only condition, fact generation in HKGs depends on both the local query and the global HKG structure. \ours is trained to capture these dependencies via a bi-level noising strategy, which provides a diverse set of conditions by simultaneously varying both the query, i.e., the intra-fact level, and the structure of the observed HKG, i.e., the inter-fact level.

\paragraph{Bi-level Noising Strategy} Bi-level noising simulates diverse conditional generation scenarios by varying both the observed structure and the query during training. First, to inject noise at inter-fact level, the HKG is dynamically partitioned every epoch via a process termed stochastic structure sampling. Specifically, the set of training facts $\sH_\text{train}$ is split into an observed set $\sH_\text{obs}$ and a target set $\sH_\text{tgt}$, where \ours computes entity and relation representations using only $\sH_\text{obs}$. This strategy forces \ours to output probability distributions conditioned on varying sets of observed facts. Next, at the intra-fact level, queries are generated by arbitrarily masking components within the facts in $\sH_\text{tgt}$. Crucially, because $\sH_\text{tgt}$ consists of facts excluded from $\sH_\text{obs}$, \ours is trained to generate facts missing in the currently observed facts, simulating the generation of new facts.

\paragraph{Diffusion objective} The training objective of \ours is the \textit{any-order autoregressive} (AO-AR) objective, which is mathematically equivalent to the masked discrete diffusion objective~\cite{radd}. The masking distribution $\sD_\xi$ is induced by first randomly selecting the number of masked components $n_\mathrm{mask}\!\sim\!\sU(\{1,\dots,2n_\xi+3\})$ and subsequently sampling $n_\mathrm{mask}$ components without replacement from $\xi$. The objective is to minimize the negative log-likelihood of the ground-truth values for the masked components:
\begin{equation}
\small
\label{eq:loss}
\sL_{\text{AO-AR}}\!=\!\!\!\underset{\substack{\xi\sim\sH_\text{tgt}\\\sM_{\zeta}\sim\sD_{\xi}}}{\mathbb{E}}\!\left[-\!\!\!\!\sum_{(x,y)\in\sM_{\zeta}}\!\!\!\!\!\log\!P_\theta(x\!=\!y\!\mid\!\zeta, (\sV,\sR,\sH_\text{obs}))\right]
\end{equation}
where $\sM_{\zeta}$ denotes the set of pairs $(x,y)$ comprising a masked component $x$ and its ground-truth $y$ in $\zeta$. By optimizing this objective, \ours is trained to reconstruct a masked query $\zeta$ in the target set using only the facts in $\sH_\text{obs}$. As a result, \ours learns to approximate the true probability distribution of masked components conditioned on the query and the observed structure, allowing it to generate plausible and novel facts absent from the training set. Note that \ours learns the time-independent reparameterization of the absorbing discrete diffusion~\cite{radd}.

\subsection{Discriminative and Generative Inference}
\label{subsec:inf}
During inference, \ours computes the representations by utilizing all facts in $\sH_\text{train}$ (Section~\ref{subsec:layer}). Given an incomplete fact, a forward pass generates mask representations to produce probability distributions for each mask (Section~\ref{subsec:pd}). These probability distributions are utilized to perform either discriminative or generative inference.

\paragraph{Link Prediction} For discriminative link prediction (Definition~\ref{def:lp}), each query contains exactly one mask. \ours leverages the computed probability distribution to rank the candidates and identifies the most likely missing component.

\paragraph{Fact Generation} For fact generation (Definition~\ref{def:fg}), the goal is to generate a valid fact from an arbitrarily masked query via iterative decoding~\cite{radd}. At each step, probability distributions for all currently masked components are recomputed using the current state of query. A random subset of these masks is selected, which are replaced via probability-based top-$p$ sampling, where $p$ is a hyperparameter. This process continues until all masks are resolved or the maximum number of steps is reached. In the latter case, all remaining masks are sampled simultaneously. In our implementation, we employ a caching strategy that recomputes mask vectors only when a mask is replaced. This bounds the computation steps by the query length $|\zeta|$, resulting in an inference cost of $\sO(|\zeta|^2d^2(L+|\sV|+|\sR|))$.

\section{Experiments}
For experiments, three HKG benchmark datasets are used: \wdk~\cite{stare}, \wppm~\cite{gran}, and \wpp~\cite{nalp}. We follow the standard evaluation protocols~\cite{stare, nalp} for link prediction. For each dataset, \ours is trained once using Eq.~\ref{eq:loss}, and directly applied to three downstream tasks: entity prediction, relation prediction, and fact generation. In our tables, the best and second-best results are \textbf{boldfaced} and \underline{underlined}, respectively. Dataset statistics and experimental details are provided in Appendix~\ref{app:data}, and hyperparameter settings are described in Appendix~\ref{app:hp}. Additional experiments using BERT~\cite{bert} and BART~\cite{bart} are reported in Appendix~\ref{app:add_exp}.

\begin{table}[t]
\centering
\footnotesize
\caption{Relation prediction on \wdk and \wppm. Pri targets primary relations, whereas All includes qualifier relations.}
\label{tb:hkgrel}
\setlength{\tabcolsep}{0.5em}
\begin{tabular}{ccccccccc}
\toprule
& & \multicolumn{3}{c}{\wdk} & \multicolumn{3}{c}{\wppm} \\
 \cmidrule(lr){3-5} \cmidrule(lr){6-8}
&  & \mrr & \hten & \hone & \mrr & \hten & \hone \\
\midrule
\multirow{4}{*}{Pri} & \gran & - & - & - & 0.957 & 0.976 & 0.942\\
& \hahe & 0.916 & 0.964 & 0.885 & 0.957 & 0.978 & 0.941 \\
& \hdiff & \underline{0.950} & \underline{0.985} & \underline{0.924} & \underline{0.973} & \underline{0.991} & \underline{0.958} \\
\cmidrule(lr){2-8}
& \ours &  \textbf{0.963} & \textbf{0.995} & \textbf{0.940} & \textbf{0.984} & \textbf{0.999} & \textbf{0.972} \\
\midrule
\multirow{4}{*}{All} & \hahe & 0.927 & 0.969 & 0.900 & 0.958 & 0.978 & 0.942 \\
& \hdiff & \underline{0.956} & \underline{0.987} & \underline{0.934} & \underline{0.973} & \underline{0.992} & \underline{0.959} \\
\cmidrule(lr){2-8}
& \ours & \textbf{0.968} & \textbf{0.995} & \textbf{0.947} & \textbf{0.984} & \textbf{0.999} & \textbf{0.972} \\
\bottomrule
\end{tabular}
\end{table}

\begin{table*}[t]
\centering
\footnotesize
\caption{Fact Generation on \wdk (denoted by WD), \wppm (denoted by WP$^-$) and \wpp (denoted by WP). Accuracy is reported. \textit{Scratch} denotes generating facts from scratch, \textit{Targeted} denotes generating facts for a single known component, and \textit{Arbitrary Masking} denotes generating facts starting from an arbitrary number of known components. N/A denotes not applicable.}
\label{tb:fg}
\begin{tabular}{llccccccccc}
\toprule
& & \multicolumn{3}{c}{\textit{Scratch}} & \multicolumn{3}{c}{\textit{Targeted}} & \multicolumn{3}{c}{\textit{Arbitrary Masking}} \\
 \cmidrule(lr){3-5} \cmidrule(lr){6-8} \cmidrule(lr){9-11}
& LLM & WD & WP$^-$ & WP & WD & WP$^-$ & WP & WD & WP$^-$ & WP \\
\midrule
\textit{Iterative Prediction} & N/A & 0.262 & 0.273 & 0.201 & 0.270 & 0.367 & 0.343 & 0.447 & 0.395 & 0.453\\
\midrule
\textit{Gibbs Sampling} & N/A & 0.299 & 0.369 & 0.230 & 0.199 & 0.291 & 0.270 & 0.390 & 0.418 & 0.397\\
 \midrule
\multirow{2}{*}{\textit{Re-ranking}} &GPT 5.2 & 0.215 & 0.229 & 0.233 & 0.112 & 0.176 & 0.192 & 0.133 & 0.177 & 0.157\\
 &Gemini 3 Pro & 0.304 & 0.288 & 0.278 & 0.256 & 0.305 & 0.293 & 0.295 & 0.314 & 0.313\\
 \midrule
\multirow{2}{*}{\textit{Neighbor Sets}} &GPT 5.2 & 0.175 & 0.124 & 0.109 & 0.397 & 0.309 & 0.355 & 0.351 & \underline{0.497} & 0.339\\
 & Gemini 3 Pro & \underline{0.351} & 0.198 & 0.204 & 0.445 & \underline{0.394} & 0.361 & \underline{0.488} & 0.475 & 0.409\\
  \midrule
\multirow{2}{*}{\textit{Few-shot Facts}} & GPT 5.2 & 0.290 & 0.233 & 0.260 & 0.222 & 0.232 & 0.238 & 0.312 & 0.340 & 0.324\\
 & Gemini 3 Pro & 0.326 & 0.307 & \underline{0.326} & 0.281 & 0.252 & 0.255 & 0.311 & 0.337 & 0.280\\
  \midrule
\multirow{2}{*}{\textit{Random Facts}} & GPT 5.2 & 0.298 & 0.212 & 0.301 & 0.463 & 0.344 & 0.351 & 0.405 & 0.458 & 0.432 \\
 & Gemini 3 Pro & 0.317 & \underline{0.343} & 0.302 & \underline{0.482} & 0.333 & \underline{0.396} & \textbf{0.604} & 0.446 & \underline{0.469}\\
 \midrule
\multirow{2}{*}{\textit{Autoregressive}} & GPT 5.2 & 0.248 & 0.145 & 0.228 & 0.250 & 0.213 & 0.301 & 0.248 & 0.257 & 0.234\\
 & Gemini 3 Pro & 0.265 & 0.154 & 0.238 & 0.190 & 0.175 & 0.239 & 0.423 & 0.350 & 0.364\\
\midrule
\ours & N/A & \textbf{0.717} & \textbf{0.855} & \textbf{0.777} & \textbf{0.508} & \textbf{0.600} & \textbf{0.591} & \textbf{0.604} & \textbf{0.649} & \textbf{0.621} \\
\bottomrule
\end{tabular}
\end{table*}

\subsection{Link Prediction Results}
\label{subsec:lp}
We evaluate \ours on two link prediction subtasks: entity prediction and relation prediction, where the Mean Reciprocal Rank (MRR) and Hits@$k$ ($k=1, 10$) are used as evaluation metrics. For all baseline methods, we report results from their original papers. Note that the number of baselines for relation prediction is limited, as only a few methods have been evaluated on this task.

Tables~\ref{tb:hkgent} and~\ref{tb:wppent} show that \ours outperforms entity prediction baselines on \wdk and \wpp while matching \maypl~\cite{maypl} on \wppm. In relation prediction, reported in Table~\ref{tb:hkgrel} and Appendix~\ref{app:rp}, \ours outperforms all baselines across all datasets. These results highlight that \ours excels at discriminative tasks despite its generative objective. We attribute this to the nature of our training: link prediction is encompassed within the masked discrete diffusion process, and the contextual message passing captures precise dependencies by explicitly conditioning on the complementary components.

\subsection{Fact Generation Results}
\label{subsec:fg}
We define three settings for fact generation: (i) \textit{Scratch}, generating a valid fact without any known component, representing the most extreme scenario of generation from scratch; (ii) \textit{Targeted}, requiring the generation of a fact containing a single given component called a target; and (iii) \textit{Arbitrary Masking}, generating a fact from an arbitrarily masked input, covering diverse scenarios including link prediction and generation from scratch. We use 1,000 fully masked queries for \textit{Scratch}. For each of the \textit{Targeted} and \textit{Arbitrary Masking} setting, 1,000 facts in the test set for link prediction are sampled. Then, these facts are masked to retain a single component for \textit{Targeted}, whereas arbitrary number of components are masked for \textit{Arbitrary Masking}. To evaluate the validity of generated facts, we employ an LLM-as-a-judge framework~\cite{llmjudge} with GPT-5.2 and report the accuracy as the ratio of correct facts. To align with the standard filtered setting in link prediction, we employ rejection sampling, where generated facts present in the training set are discarded and re-generated. If a model fails to generate a novel fact within 10 attempts, we penalize it by marking the result as incorrect. More details of fact generation experiments are in Appendix~\ref{app:fg}.

Since existing HKG models are discriminative, they cannot be directly applied to generative tasks. Therefore, we design seven competitive baselines: two variations of discriminative HKG models and five LLM-based approaches tailored for fact generation. We employ the state-of-the-art HKG model, \maypl, as the backbone of the baselines that require an HKG model. We use GPT 5.2~\cite{gpt} and Gemini 3.0 Pro~\cite{gemz} for the LLM-based methods. Details about these baseline methods are in Appendix~\ref{app:fgbase}.

\begin{compactitem}
\item \textit{Iterative Prediction}: Missing components are randomly initialized. Subsequently, they are sequentially masked and replaced by an HKG model's prediction.
\item \textit{Gibbs Sampling:} Missing components are randomly initialized, followed by an Gibbs sampling process where each missing component is iteratively masked and replaced by resampling from the distribution predicted by an HKG model.
\item \textit{Re-ranking:} Missing components are initialized using a training fact that matches the known components of the query. Each missing component is sequentially masked, and an LLM selects the best candidate from the top-5 predictions of an HKG model.
\item \textit{Neighbor Sets:} An LLM generates each fact prompted by the neighbors of known components in the query.
\item \textit{Few-shot Facts:} An LLM generates each fact prompted by 30 training facts containing at least one known entity or relation from the query.
\item \textit{Random Facts:} An LLM is prompted to generate facts by utilizing 1,000 randomly sampled training facts.
\item \textit{Autoregressive:} An LLM generates components step-by-step, prompted to utilize training facts containing the previously predicted component for generation. \end{compactitem}

Table~\ref{tb:fg} presents the results for fact generation. Across all settings, \ours consistently achieves state-of-the-art performance, demonstrating superior generative capabilities. This stems from the design of \ours, which explicitly learns to generate plausible and new facts via contextual message passing and bi-level noising strategy. Unlike LLMs, which generates facts by modeling the likelihood of general text sequences, \ours is directly optimized to model the probability distribution of missing components conditioned on the query and the global structure of an HKG. A critical distinction also lies in the novelty of the generated facts. While it is often indistinguishable whether LLMs are performing reasoning or merely retrieving memorized information from their massive pre-training corpora, the facts used to evaluate \ours are clearly not in its training data. Thus, the accuracy reflects the capacity of \ours to infer novel and correct facts. \ours effectively performs conditional generation for incomplete facts, significantly outperforming LLMs in HKG-conditioned generation tasks.

\begin{table}[t]
\centering
\footnotesize
\caption{Fact Generation from scratch on \wppm. We report the Valid \& Novel Rate (V\&N Rate) and the expected number of generation attempts needed per valid and novel fact ($\EE[\text{Gen.}]$). \ours achieves the highest V\&N Rate and the lowest $\EE[\text{Gen.}]$.}
\label{tb:rej}
\setlength{\tabcolsep}{0.25em}
\begin{tabular}{llcc}
\toprule
& LLM & V\&N Rate($\uparrow$) & $\EE[\text{Gen.}]$($\downarrow$) \\
\midrule
\textit{Iterative Prediction} & N/A & 0.204 & 4.90\\
\midrule
\textit{Gibbs Sampling} & N/A & 0.184 & 5.44 \\
 \midrule
\multirow{2}{*}{\textit{Re-ranking}} &GPT 5.2 & 0.157 & 6.36\\
 &Gemini 3.0 Pro & 0.155 & 6.45 \\
 \midrule
\multirow{2}{*}{\textit{Neighbor Sets}} &GPT 5.2 & 0.110 & 9.11 \\
 & Gemini 3.0 Pro & 0.136 & 7.37\\
  \midrule
\multirow{2}{*}{\textit{Few-shot Facts}} & GPT 5.2 & 0.153 & 6.54\\
 & Gemini 3.0 Pro & 0.242 & 4.13\\
  \midrule
\multirow{2}{*}{\textit{Random Facts}} & GPT 5.2 & 0.036 & 27.58 \\
 & Gemini 3.0 Pro & 0.104 & 9.65 \\
 \midrule
\multirow{2}{*}{\textit{Autoregressive}} & GPT 5.2 & 0.089 & 11.27 \\
 & Gemini 3.0 Pro & 0.091 & 10.99 \\
\midrule
\ours & N/A & \textbf{0.351} & \textbf{2.85}\\
\bottomrule
\end{tabular}
\end{table}

\paragraph{Additional Metrics} To comprehensively measure both accuracy and novelty, we introduce two supplementary metrics: the Valid \& Novel Rate (V\&N Rate), the proportion of correct and novel facts, and the expected number of generation attempts per valid and novel fact ($\EE[\text{Gen.}]$). A higher V\&N Rate and a lower $\EE[\text{Gen.}]$ indicate superior performance. Table~\ref{tb:rej} reports these metrics on \wppm under the Scratch setting. \ours maintains the best performance, confirming its capability to generate novel, accurate facts.

\subsection{LLM-as-a-Judge Verification}
LLM-as-a-Judge is a common alternative to human evaluation~\cite{llmjudge}. To validate our GPT-5.2 judge, we conduct two supplementary evaluations on \wppm under the Scratch setting:
\begin{compactitem}
\item \textit{Multi-Judge Consensus}: We employ two additional LLM judges, Gemini 3.1 Pro~\cite{gemo} and Grok 4.20 Reasoning~\cite{grok}. A generated fact is considered correct only if GPT 5.2, Gemini 3.1 Pro, and Grok 4.20 Reasoning reach a unanimous agreement.
\item \textit{Human Evaluation}: We sample a 10\% subset of the generated facts and engage three human annotators. A fact is considered correct only upon unanimous agreement among all three humans.
\end{compactitem}


\begin{table}[t]
\centering
\footnotesize
\caption{Comparison of accuracy between the GPT-5.2 Judge and Multi-Judge Consensus (Multi) on the full set, and a comparison between the GPT-5.2 Judge and Human evaluation on a 10\% subset. All evaluations are conducted for fact generation from scratch on \wppm. The trend remains consistent across different evaluators. GPT and Gemini denote GPT 5.2 and Gemini 3.0 Pro.}
\label{tb:llmjudge_val}
\setlength{\tabcolsep}{0.25em}
\begin{tabular}{llcccc}
\toprule
& & \multicolumn{2}{c}{\textit{Full set}} & \multicolumn{2}{c}{\textit{10\% Subset}}\\
 \cmidrule(lr){3-4} \cmidrule(lr){5-6}
& LLM & GPT-5.2 & Multi & GPT-5.2 & Human \\
\midrule
\textit{Iterative Prediction} & N/A & 0.273 & 0.254 & 0.24 & 0.25 \\
\midrule
\textit{Gibbs Sampling} & N/A & 0.369 & 0.343 & 0.33 & 0.25\\
 \midrule
\multirow{2}{*}{\textit{Re-ranking}} &GPT& 0.229 & 0.163 & 0.28 & 0.22\\
 &Gemini & 0.288 & 0.263 & 0.32 & 0.32\\
 \midrule
\multirow{2}{*}{\textit{Neighbor Sets}} &GPT& 0.124 & 0.107 & 0.15 & 0.11\\
 & Gemini & 0.198 & 0.160 & 0.17 & 0.17\\
  \midrule
\multirow{2}{*}{\textit{Few-shot Facts}} & GPT & 0.233 & 0.217 & 0.21 & 0.21\\
 & Gemini & 0.307 & 0.273 & 0.29 & 0.31\\
  \midrule
\multirow{2}{*}{\textit{Random Facts}} & GPT & 0.212 & 0.184 & 0.20 & 0.18 \\
 & Gemini & 0.343 & 0.309 & 0.42 & 0.38\\
 \midrule
\multirow{2}{*}{\textit{Autoregressive}} & GPT & 0.145 & 0.126 & 0.18 & 0.14\\
 & Gemini & 0.154 & 0.141 & 0.18 & 0.18\\
\midrule
\ours & N/A & \textbf{0.855} & \textbf{0.821} & \textbf{0.86} & \textbf{0.83}\\
\bottomrule
\end{tabular}
\end{table}

Table~\ref{tb:llmjudge_val} details the evaluation results, demonstrating that while strict unanimous criteria slightly lower the absolute scores, the overall model rankings remain highly stable. To formally verify this alignment, we calculate the Pearson correlation coefficients, where 1 implies an identical trend. The results reveal a near-perfect positive correlation: 0.997 between GPT-5.2 and Multi-Judge on the full set, and 0.987 between GPT-5.2 and Human on the 10\% subset. These coefficients confirm that the scoring trends are almost identical, proving that our evaluation remains highly consistent across diverse LLMs and human judgment.

\begin{table*}[t]
\centering
\footnotesize
\caption{Qualitative examples of fact generation by \ours on \wdk. Generated facts with zero, one, and two qualifiers for the head entity ``Toy Story" and tail entity ``A. Irving" are presented. Additionally, three facts generated from scratch are shown.}
\label{tb:case:kg}
\setlength{\tabcolsep}{0.14em}
\begin{tabular}{cl}
\toprule
Query & Generation \\
\midrule
\multirow{3}{*}{$\big((\text{Toy Story}, \texttt{?}, \texttt{?}), \{(\texttt{?},\texttt{?})\}_{i=1}^{n_\xi}\big)$} & $\big((\text{Toy Story}, \text{genre}, \text{fantasy film}),\{\}\big)$ \\
\noalign{\vskip\aboverulesep}\cdashline{2-2}\noalign{\vskip\belowrulesep}
& \big((Toy Story, voice actor, T. Allen), \{(as, Buzz Lightyear)\}\big)\\
\noalign{\vskip\aboverulesep}\cdashline{2-2}\noalign{\vskip\belowrulesep}
& \big((Toy Story, nominated for, Oscars Best Score), \{(subject of, 68th Oscars), (nominee, R. Newman)\}\big)\\
\midrule
\multirow{3}{*}{\big((\texttt{?}, \texttt{?}, A. Irving), $\{(\texttt{?},\texttt{?})\}_{i=1}^{n_\xi}$\big)} & \big((She's Having a Baby, cast member, A. Irving),\{\}\big) \\
\noalign{\vskip\aboverulesep}\cdashline{2-2}\noalign{\vskip\belowrulesep}
& \big((SAGA Performance Cast, winner, A. Irving), \{(for work, Traffic)\}\big)\\
\noalign{\vskip\aboverulesep}\cdashline{2-2}\noalign{\vskip\belowrulesep}
& \big((Who Framed Roger Rabbit, voice actor, A. Irving), \{(as, Jessica Rabbit), (role, singer)\}\big)\\
\midrule
\multirow{3}{*}{\big((?,?,?), \{(?,?)\}\big)} & \big((Star Wars: Episode IV, distributed by, InterCom), \{(country, Hungary)\}\big) \\
\noalign{\vskip\aboverulesep}\cdashline{2-2}\noalign{\vskip\belowrulesep}
& \big((L. Messi, participant in, 2008 Champions League), \{(team, F.C. Barcelona)\}\big) \\
\noalign{\vskip\aboverulesep}\cdashline{2-2}\noalign{\vskip\belowrulesep}
& \big((Hugo Award for Best Novel, winner, C. J. Cherryh), \{(for work, Cyteen)\}\big) \\
\bottomrule
\end{tabular}
\end{table*}

\begin{table}[t]
\centering
\footnotesize
\caption{Ablation studies of \ours on \wdk and \wppm. We report \mrr for Entity Prediction on all positions (LP) and Accuracy for Fact Generation from scratch (FG).}
\label{tb:abl}
\begin{tabular}{clcccc}
\toprule
& & \multicolumn{2}{c}{\wdk} & \multicolumn{2}{c}{\wppm}\\
 \cmidrule(lr){3-4} \cmidrule(lr){5-6}
& & LP & FG & LP & FG\\
\midrule
(i) & w/o Context Msg. & 0.408 & 0.552 & 0.507 & 0.578 \\
(ii) & w/o Str. Samp. & 0.296 & 0.545 & 0.409 & 0.681 \\
(iii) & w/o Attention & 0.408 & 0.673 & 0.466 & 0.779 \\
(iv) & w/ Individual Init. & 0.272 & 0.466 & 0.390 & 0.575\\
(v) & w/ LP Loss & 0.415 & 0.038 & 0.516 & 0.002\\
(vi) & w/ Cos. Sim. & \textbf{0.419} & 0.711 & 0.497 & 0.825 \\
\midrule
& \ours &  \textbf{0.419} & \textbf{0.717} & \textbf{0.522} & \textbf{0.855}\\
\bottomrule
\end{tabular}
\end{table}

\subsection{Ablation Studies}

Table~\ref{tb:abl} presents the ablation studies of \ours on \wdk and \wppm. \textbf{(i)} Removing context messages (Eq.~\ref{eq:cme},~\ref{eq:cmr}) and relying solely on fact representations~(Eq.\ref{eq:fact}) to update entities and relations (Section~\ref{subsec:layer}) leads to substantial degradation, confirming that excluding self information and aggregating surrounding contexts is crucial, especially for fact generation. \textbf{(ii)} Disabling stochastic structure sampling (Section~\ref{subsec:tr}) degrades performance, indicating that varying the observed set during training is essential. \textbf{(iii)} Replacing attention in contextual message passing with mean pooling(Section~\ref{subsec:layer}) leads to worse results, underscoring the importance of adaptively weighting facts. \textbf{(iv)} Using individual initializations for every entity and relation instead of our shared initialization (Section~\ref{subsec:layer}) reduces performance, suggesting that our approach outperforms the use of unique identifiers. \textbf{(v)} Replacing the AO-AR loss with the cross-entropy loss for link prediction yields near-zero generation accuracy. This proves that our objective unifies both capabilities without compromising discriminative performance. \textbf{(vi)} Replacing the dot product with cosine similarity in Eq.~\ref{eq:pd} degrades performance, highlighting the significance of the magnitude in the representation space. We also provide the runtime-quality tradeoff for denoising steps in Appendix~\ref{app:step}.

\subsection{Qualitative Analysis of \ours}
\label{subsec:qual}
Table~\ref{tb:case:kg} presents a qualitative analysis, illustrating facts generated by \ours for given queries. First, we examine how the generated output varies with the number of qualifiers $n_\xi$ when the head or tail entity is fixed. For instance, given the head entity ``Toy Story" and $n_\xi\!=\!2$, \ours generates a fact representing the knowledge that the movie Toy Story was nominated for the Oscars for Best Score at the 68th Oscars, with the nominee R. Newman. When generating from scratch with a single qualifier pair, \ours produces a diverse set of facts across various domains such as film, sports, and literature. These results demonstrate that \ours generates both plausible and novel facts for a given query. Appendix~\ref{app:cs} provides further insights, including the scratch generation results across varying lengths, a step-by-step analysis of the generation process, and case studies on the top-3 similar entities or relations for a target.

\section{Conclusion and Future Work}
We introduce \textit{Fact Generation}, a novel task that advances beyond the traditional fill-in-the-blank link prediction in the HKG domain by aligning more closely with real-world scenarios where multiple components of a fact may be missing. We propose \ours, the first HKG representation learning framework equipped with generative capabilities via a masked discrete diffusion objective. By employing contextual message passing to model conditional dependencies within a fact and stochastic structure sampling to capture inter-fact relationships, \ours effectively learns the joint conditional distribution of masked components. Empirical results demonstrate the superiority of our approach: \ours not only establishes a new state-of-the-art link prediction performance but also significantly outperforms LLM-based baselines in fact generation. For future work, we plan to extend our framework to inductive scenarios, applying it to inference graphs containing unseen entities and relations. Also, we intend to explore \ours's generative capabilities across a wide range of applications that require structured reasoning and the discovery of new knowledge and facts.

\section*{Impact Statement}
This paper introduces a framework for fact generation on HKGs. By enabling models to generate structurally valid and plausible facts beyond simple link prediction, our work has the potential to significantly accelerate the automated completion of large-scale knowledge bases in critical domains such as biomedicine, finance, and enterprise knowledge management. Unlike LLMs that may generate structurally inconsistent facts, our approach ensures adherence to the rigorous topology of HKGs, potentially offering a more reliable source for structured knowledge reasoning.

However, the advanced generative capabilities of \ours also present potential risks regarding the dissemination of misinformation. Since the model is proficient at generating structurally coherent and plausible facts even from scratch, it could potentially be misused to automate the creation of fabricated knowledge. Malicious actors might leverage this capability to inject high-fidelity false information into broader information ecosystems, thereby corrupting downstream applications that rely on the factual integrity of the ecosystem. Furthermore, the model's ability to infer missing components without rigid constraints necessitates careful monitoring to prevent the unintentional reinforcement of biases present in the training data. Therefore, the deployment of such generative models requires robust verification mechanisms and human supervision to distinguish between genuine knowledge discovery and malicious or manipulated content.

\section*{Acknowledgements}
This work was partly supported by the National Research Foundation of Korea(NRF) grant funded by the Korean government(MSIT) (70\% from RS-2025-00559066), Institute of Information \& communications Technology Planning \& Evaluation(IITP) grant funded by the Korea government(MSIT) (24\% from RS-2022-II220369, (Part 4) Development of AI Technology to support Expert Decision-making that can Explain the Reasons/Grounds for Judgment Results based on Expert Knowledge, 1\% from RS-2025-25442149, LG AI STAR Talent Development Program for Leading Large-Scale Generative AI Models in the Physical AI Domain), and the InnoCORE program of the Ministry of Science and ICT (5\% from AI Meta-Scientist, N10260110).


\bibliography{diffhkg}
\bibliographystyle{icml2026}

\newpage
\appendix
\onecolumn
\section{Implementation Details of \ours}
\label{app:imp}

We implement \ours based on the Pre-LN architecture~\cite{preln} to ensure training stability, utilizing the Gaussian Error Linear Unit (GELU)~\cite{gelu} for non-linear activation. A scaled dot-product is employed to compute the final candidate probabilities as formulated in Eq.~\ref{eq:pd}. Bias terms are included in all linear projections, and dropout is employed. Additionally, we set $\vz_\mathrm{ENT}=\vx_\mathrm{ENT}$ and $\vz_\mathrm{REL}=\vx_\mathrm{REL}$ for \wpp and \wppm. In terms of regularization, we exclude bias terms and LayerNorm parameters from weight decay following common practice. For \wpp and \wppm, weight decay is explicitly applied to the LayerNorm preceding the MLP layer. Our code and datasets are available at \url{https://github.com/bdi-lab/KREPE}.

We use Python 3.11 and PyTorch 2.6.0 with CUDA 11.8. \ours is optimized via AdamW~\cite{adamw}. We utilize an NVIDIA GeForce RTX 3090 for \wdk and an NVIDIA RTX A6000 for \wpp and \wppm.

\section{Details on the HKG Datasets}
\label{app:data}
\begin{table}[ht]
\centering
\small
\caption{Statistics of HKG datasets.}
\label{tb:stat}
\begin{tabular}{cccccc}
\toprule
 & $|\sV|$ & $|\sR|$ & $|\sH_\text{train}|$ & $|\sH_\text{valid}|$ & $|\sH_\text{test}|$ \\
 \midrule
\wdk & 47,155 & 531 & 166,435 & 23,913 & 46,159 \\
\wppm & 34,825 & 178 & 294,439 & 37,715 & 37,712 \\
\wpp & 47,765 & 193 & 305,725 & 38,223 & 38,281 \\
\bottomrule
\end{tabular}
\end{table}

In our experiments, we utilize three standard benchmarks: \wdk~\cite{stare}, \wppm~\cite{gran}, and \wpp~\cite{nalp}. \wdk is an HKG dataset derived from Wikidata by extracting facts that contain entities that correspond to those in FB15K-237~\cite{fb}. \wpp is another HKG dataset from Wikidata, collected by filtering facts involving humans, while \wppm is a modified version of \wpp constructed by removing literals. The detailed statistics for each dataset are summarized in Table~\ref{tb:stat}. $\sH_\text{train}$, $\sH_\text{valid}$, and $\sH_\text{test}$ denote the training, validation, and test set, respectively.

On \wdk and \wppm, we select optimal hyperparameters using the validation set and report final results after retraining \ours on the combined training and validation sets, by following the common practice for these datasets~\cite{stare,gran}. For \wpp, we directly utilized the model trained on the training set with the best validation performance. Following \maypl~\cite{maypl}, we transfer the hyperparameter configuration from \wppm to \wpp, tuning only the number of training epochs based on validation performance.

\section{Hyperparameter Settings of \ours}
\label{app:hp}

\begin{table}[t]
\centering
\small
\caption{Best hyperparameter settings of \ours for \wdk, \wppm, and \wpp.}
\label{tb:thp}
\begin{tabular}{ccccccccccccccccc}
\toprule
 & best epoch & $\eta_\mathrm{max}$ & $\eta_{\mathrm{min}}$ & $d$ & $L$ & $n_\mathrm{ENT}$ & $n_\mathrm{REL}$ & $B$\\
 \midrule
\wdk & 1950 & 0.001 & 0.00001 & 128 & 16 & 4 & 4 & 2048 \\
\wppm & 2000 & 0.0004 & 0.00001 & 256 & 12 & 8 & 32 & 4096 \\
\wpp & 1750 & 0.0004 & 0.00001 & 256 & 12 & 8 & 32 & 4096 \\
\bottomrule
\end{tabular}
\end{table}

We perform a grid search to determine the optimal hyperparameters for each dataset. Table~\ref{tb:thp} reports the best hyperparameter configurations for \ours. We employ a cosine learning rate scheduler~\cite{coslin} over a total of 2,000 epochs, incorporating a linear warmup phase for the first 200 epochs. The peak learning rate and the minimum learning rate is denoted by $\eta_\mathrm{max}$ and $\eta_\mathrm{min}$, respectively. Validation is performed every 50 epochs to identify the best epoch. Additionally, we use a weight decay coefficient of $0.01$ with AdamW for \wdk and $0.1$ for \wpp and \wppm. For the stochastic structure sampling proposed in Section~\ref{subsec:tr}, the probability of sampling a fact from $\sH_\text{train}$ to form the observed set $\sH_\text{obs}$ is fixed at 0.7. To stabilize training, we apply gradient clipping with a maximum norm of 1.0. The number of attention heads is specified separately for entities ($n_\mathrm{ENT}$) and relations ($n_\mathrm{REL}$). $B$ refers to the batch size. We use a dropout rate of 0.1 across all datasets.

\paragraph{Sampling Parameters}
As described in Section~\ref{subsec:inf}, we employ a probability-based top-$p$ sampling for fact generation, where $\delta_\mathrm{ENT}$ and $\delta_\mathrm{REL}$ denote the probability thresholds for entity and relation sampling, respectively. Following~\cite{radd}, we adopt a log-linear noise schedule over 1,000 reconstruction steps to ensure that the AO-AR objective is equivalent to the absorbing diffusion loss~\cite{radd}. Additionally, we apply temperature scaling to control the stochasticity of generation, denoted by $\tau_\mathrm{ENT}$ and $\tau_\mathrm{REL}$. For \wdk, the \textit{Scratch} setting uses thresholds of $\delta_\mathrm{ENT}=0.15$ and $\delta_\mathrm{REL}=0.05$ with unit temperature ($\tau_\mathrm{ENT}=\tau_\mathrm{REL}=1$). For the \textit{Targeted} and \textit{Arbitrary Masking} settings, we employ $\delta_\mathrm{ENT}=0.7$, $\delta_\mathrm{REL}=0.5$, $\tau_\mathrm{ENT}=0.55$, and $\tau_\mathrm{REL}=0.5$. For \wppm, we use $\delta_\mathrm{ENT}=\delta_\mathrm{REL}=0.2$ and $\tau_\mathrm{ENT}=\tau_\mathrm{REL}=1$ across all settings. For \wpp, we use $\delta_\mathrm{ENT}=0.2$, $\delta_\mathrm{REL}=0.1$, and $\tau_\mathrm{ENT}=\tau_\mathrm{REL}=1$ across all settings.

\section{Relation Prediction Results for \wpp}
\label{app:rp}

\begin{table}[t]
\centering
\footnotesize
\caption{Relation prediction on all relations in facts in \wpp.}
\label{tb:wpprel}
\begin{tabular}{ccccc}
\toprule
 & \mrr & \hten & \hone \\
\midrule
\nalp & 0.735 & \underline{0.938} & 0.595 \\
\neuinf & 0.765 & 0.897 & 0.686 \\
\tnalp & \underline{0.790} & 0.927 & \underline{0.713} \\
\midrule
\ours & \textbf{0.985} & \textbf{0.999} & \textbf{0.972} \\
\bottomrule
\end{tabular}
\end{table}

Table~\ref{tb:wpprel} reports the relation prediction results on \wpp. Consistent with the results on \wdk and \wppm, \ours significantly outperforms all existing methods. As noted in Section~\ref{subsec:lp}, the baseline comparison is limited to a subset of those used for entity prediction, as relation prediction is less frequently evaluated in the literature.

\section{Additional Baseline Results for \wppm}
\label{app:add_exp}

\begin{wraptable}{r}{0.48\textwidth}
\centering
\vspace{-15pt}
\footnotesize
\caption{End-to-end training results using BERT and BART architecture for \wppm. We report the \mrr for link prediction and the accuracy for fact generation from scratch.}
\label{tb:wppm_brt}
\begin{tabular}{ccc}
\toprule
 & Link Prediction & Fact Generation \\
\midrule
BERT & 0.503 & 0.098 \\
BART & 0.477 & 0.181 \\
\midrule
\ours & \textbf{0.522} & \textbf{0.855} \\
\bottomrule
\end{tabular}
\end{wraptable}

Table~\ref{tb:wppm_brt} reports the performance of BERT~\cite{bert}, a masked language model, and BART~\cite{bart}, an autoregressive transformer, on \wppm. We serialized the facts into sequences and performed end-to-end training for these architectures. We report the \mrr for link prediction (LP) and the accuracy for fact generation from scratch (FG). The results show that \ours significantly outperforms these baselines in both tasks.

\section{Runtime-Quality Tradeoff in Denoising Steps of \ours}
\label{app:step}

\begin{wraptable}{r}{0.46\textwidth}
\centering
\vspace{-10pt}
\footnotesize
\caption{Runtime-quality tradeoffs for varying denoising steps of \ours on \wppm for fact generation from scratch.}
\label{tb:step}
\begin{tabular}{lcc}
\toprule
Steps & Time(sec/query) & Accuracy \\
\midrule
50 & 0.530 & 0.671 \\
100 & 0.550 & 0.684 \\
250 & 0.647 & 0.708 \\
500 & 0.757 & 0.768 \\
1000 & 0.910 & 0.855 \\
\bottomrule
\end{tabular}
\end{wraptable}

We provide the runtime-quality tradeoffs for fewer denoising steps on \wppm for fact generation from scratch. We also report the time required to generate a single query with a batch size of 1. Table~\ref{tb:step} shows that reducing steps decreases both runtime and accuracy. Yet, even with just 50 steps, the accuracy of \ours (0.671) is nearly 2x higher than the best baseline, Gibbs Sampling (0.369), while being over 5$\times$ faster (0.53 sec vs. 2.81 sec). We note that \ours supports efficient batch processing; for instance, generating a batch of 1,000 queries takes only 79 seconds, yielding a latency of $\sim$0.08 sec per query.

\begin{table}
\centering
\small
\caption{Distribution of fact lengths in each dataset for the Scratch setting.}
\label{tb:fglen}
\begin{tabular}{ccccccc}
\toprule
& 3 & 5 & 7 & 9 & 11 & 13 \\
 \midrule
\wdk & 864 & 89 & 40 & 4 & 2 & 1\\
\wppm & 974 & 20 & 5 & 1 & 0 & 0 \\
\wpp & 884 & 67 & 40 & 7 & 2 & 0 \\
\bottomrule
\end{tabular}
\end{table}

\begin{figure}
\begin{tcolorbox}[colback=backcolor, colframe=framecolor, title=Fact Verification Prompt Template]
\textbf{\textcolor{usercolor}{User:}} \\
You are an expert at verifying hyper-relational facts in a hyper-relational knowledge graph.
Given a hyper-relational fact, you need to verify whether the fact is (1) semantically valid and (2) factually true.
If you have understood your responsibility, respond ``Yes". Otherwise, respond ``No". Do not output anything except ``Yes" or ``No".

\vspace{2mm} 
\textbf{\textcolor{modelcolor}{Model:}} \\
Yes

\tcbline

\textbf{\textcolor{usercolor}{User:}} \\
The goal statement is: ``\textcolor{examplecolor}{\texttt{\{{fact\}}}}''.
To verify the goal statement, you need to refer to other examples that may be similar or related to it.
Some of the given examples are similar to the goal statement. You should draw analogies from them to understand the potential meaning of the goal statement.
Other provided facts contain supplementary information; capture this extra information and mine potential relationships among them to support the verification.
Please carefully read, analyze, and reflect on these examples. Identify the reasoning patterns demonstrated in these examples and retain any information that may help your verification task.
While I provide examples, please remain silent until I ask you to respond.

\vspace{2mm} 
\textbf{\textcolor{modelcolor}{Model:}} \\
I have received your instructions and the goal statement.

\tcbline

\textbf{\textcolor{usercolor}{User:}} \\
Examples used for analogy: \\
Verify the fact ``\textcolor{examplecolor}{\texttt{\{{analogy\_fact\#1\}}}}''. The fact is semantically valid and factually true, so the verification result is Yes.\\
$\cdots$\\

Examples used to supplement information: \\
Verify the fact ``\textcolor{examplecolor}{\texttt{\{{supplement\_fact\#1\}}}}''. The fact is semantically valid and factually true, so the verification result is Yes.\\
$\cdots$\\

Keep thinking, but DO NOT give me any feedback.

\vspace{2mm} 
\textbf{\textcolor{modelcolor}{Model:}} \\
I have analyzed the examples provided and will keep thinking without giving feedback.

\tcbline

\textbf{\textcolor{usercolor}{User:}} \\
The goal is to verify the fact ``\textcolor{examplecolor}{\texttt{\{{fact\}}}}''. Based on the previous examples and your own knowledge, provide a verification result.
\begin{compactitem}
    \item If the statement is semantically valid and factually true, respond ``Yes''.
    \item If the statement is semantically valid but factually false, respond ``Half''.
    \item If the statement is semantically invalid, respond ``No''.
\end{compactitem}
Your answer must be consistent with all provided examples, even if it conflicts with standard linguistic definitions. \\

\textbf{DO NOT OUTPUT ANYTHING ELSE.}

\vspace{2mm} 
\textbf{\textcolor{modelcolor}{Model:}} \\
Yes/Half/No

\end{tcolorbox}
\captionof{figure}{The prompt used to evaluate the correctness of generated facts.}
\label{fig:eval}
\end{figure}

\section{Dataset Construction and Evaluation Protocols for Fact Generation}
\label{app:fg}

\paragraph{Dataset Construction} In Section~\ref{subsec:fg}, we defined three settings for fact generation: \textit{Scratch}, \textit{Targeted}, and \textit{Arbitrary Masking}. Since the fact generation task requires queries to initiate the generation process, we construct 1,000 queries for each setting. We define the ``base set" as the set of facts utilized to train the model for evaluation, which comprises the union of the training and validation sets for \wdk and \wppm, and strictly the training set for \wpp, following the standard conventions for these datasets~\cite{stare,gran,nalp}. For the \textit{Scratch} setting, target lengths were sampled to match the distribution of fact lengths in the base sets. Table~\ref{tb:fglen} shows the number of facts per length for each dataset. For the \textit{Targeted} setting, we sampled 1,000 facts from the test set and masked all but one component (500 kept an entity, 500 kept a relation). For the \textit{Arbitrary Masking} setting, we sampled 1,000 facts from the test set and randomly masked an arbitrary number of components ($\ge 1$).

\paragraph{Evaluation Protocol} \label{app:fg_eval} To ensure rigorous evaluation, we filtered out facts that are already present in the base sets or duplicates generated from the same query. The validity and the correctness of the generated facts were assessed as follows:
\begin{compactitem}
\item\textbf{Incorrect:} Facts containing invalid entities or relations, or those that do not match the length of the query.
\item \textbf{Correct:} Facts that exactly match the ground-truth. This includes matches in the test set for \wdk and \wppm, and matches in either the validation or test sets for \wpp.
\item \textbf{LLM Verification:} For structurally valid but non-matching facts with the test set, GPT-5.2 acts as a judge. To provide information about the entities and relations within the generated fact, the judge is given a reference example fact for each component. We refer to the example facts for the relations as `analogy facts' and the example facts for the entities as `supplement facts', following~\cite{kicgpt}. Figure~\ref{fig:eval} shows the prompt, which is adapted from~\cite{kicgpt}. The judge classifies a fact as \textbf{Yes} if it is factually correct, \textbf{Half} if it is semantically valid but factually wrong, or \textbf{No} otherwise.
\end{compactitem}
Accuracy is reported as the proportion of \textbf{Yes} labels, where \textbf{Correct} also counts as \textbf{Yes}.

\section{Additional Qualitative Analysis of \ours}
\label{app:cs}

\begin{table}[t]
\centering
\footnotesize
\caption{Examples of facts generated by \ours on \wdk, with zero, one, and two qualifiers.}
\label{tb:case:scrlen}
\begin{tabular}{cl}
\toprule
Query & Generation \\
\midrule
\multirow{3}{*}{\big((?,?,?), $\{(?,?)\}_{i=1}^{n_\xi}$\big)} & \big((Benjamin Franklin, occupation, writer), \{\}\big) \\
\noalign{\vskip\aboverulesep}\cdashline{2-2}\noalign{\vskip\belowrulesep}
& \big((L. Messi, participant in, 2008 Champions League), \{(team, F.C. Barcelona)\}\big) \\
\noalign{\vskip\aboverulesep}\cdashline{2-2}\noalign{\vskip\belowrulesep}
& \big((Gigi, awarded, Oscar Best Adapted Screenplay), \{(subject of, 31st Oscars), (winner, A. J. Lerner)\}\big) \\
\bottomrule
\end{tabular}

\bigskip

\footnotesize
\caption{A step-by-step generation trajectory for the fact \big((Gigi, awarded, Oscar Best Adapted Screenplay), \{(subject of, 31st Oscars), (winner, A. J. Lerner)\}\big) by \ours on \wdk.}
\begin{tabular}{l}
\toprule
 \big((\texttt{?}, \texttt{?}, \texttt{?}), \{(\texttt{?}, \texttt{?}), (\texttt{?}, \texttt{?})\}\big) \\
\noalign{\vskip\aboverulesep}\cdashline{1-1}\noalign{\vskip\belowrulesep}
 \big((\texttt{?}, \texttt{?}, \texttt{?}), \{(\texttt{?}, 31st Oscars), (\texttt{?}, \texttt{?})\}\big) \\
\noalign{\vskip\aboverulesep}\cdashline{1-1}\noalign{\vskip\belowrulesep}
 \big((\texttt{?}, awarded, \texttt{?}), \{(\texttt{?}, 31st Oscars), (\texttt{?}, \texttt{?})\}\big) \\
\noalign{\vskip\aboverulesep}\cdashline{1-1}\noalign{\vskip\belowrulesep}
 \big((Gigi, awarded, \texttt{?}), \{(\texttt{?}, 31st Oscars), (\texttt{?}, \texttt{?})\}\big) \\
\noalign{\vskip\aboverulesep}\cdashline{1-1}\noalign{\vskip\belowrulesep}
 \big((Gigi, awarded, \texttt{?}), \{(subject of, 31st Oscars), (\texttt{?}, \texttt{?})\}\big) \\
\noalign{\vskip\aboverulesep}\cdashline{1-1}\noalign{\vskip\belowrulesep}
 \big((Gigi, awarded, \texttt{?}), \{(subject of, 31st Oscars), (winner, \texttt{?})\}\big) \\
\noalign{\vskip\aboverulesep}\cdashline{1-1}\noalign{\vskip\belowrulesep}
 \big((Gigi, awarded, \texttt{?}), \{(subject of, 31st Oscars), (winner, A. J. Lerner)\}\big) \\
\noalign{\vskip\aboverulesep}\cdashline{1-1}\noalign{\vskip\belowrulesep}
\big((Gigi, awarded, Oscar Best Adapted Screenplay), \{(subject of, 31st Oscars), (winner, A. J. Lerner)\}\big) \\
\bottomrule
\end{tabular}
\label{tb:case:step}

\bigskip

\parbox{0.48\linewidth}{
\centering
\footnotesize
\caption{Top 3 similar entities for target entity ``Utah" based on the final representations from \ours on \wdk.}
\label{tb:case:ent}
\begin{tabular}{cccc}
\toprule
Target & & Similar Entities \\
\midrule
\multirow{3}{*}{Utah} & \#1 & Arizona\\
& \#2 & New Mexico\\
& \#3 & Colorado\\
\bottomrule
\end{tabular}
}
\hfill
\parbox{0.48\linewidth}{
\centering
\footnotesize
\caption{Top 3 similar relations for target relation ``relative" based on the final representations from \ours on \wdk.}
\label{tb:case:rel}
\begin{tabular}{cccc}
\toprule
Target & & Similar Relations \\
\midrule
\multirow{3}{*}{relative} & \#1 & sibling\\
& \#2 & family\\
& \#3 & child\\
\bottomrule
\end{tabular}
}

\end{table}

Table~\ref{tb:case:scrlen} presents facts generated by \ours on \wdk when varying numbers of qualifiers are specified. Consistent with the qualitative analysis in Section~\ref{subsec:qual}, \ours produces correct and diverse facts for a given query with varying lengths.

Table~\ref{tb:case:step} illustrates the step-by-step generation process of \ours for a fact with two qualifiers. The process begins by predicting a qualifier entity, ``31st Oscars'', which establishes the topical context. This is followed by the primary relation, ``awarded'', narrowing the scope of the fact to awards. Subsequently, the head entity is generated, further resolving the fact to the movie ``Gigi''. Next, the model predicts the qualifier relation ``subject of'', linking it to the previously generated qualifier entity ``31st Oscars''. The second qualifier relation is then generated, followed by its associated entity. Finally, the tail entity is predicted to complete the fact.

Tables~\ref{tb:case:ent} and \ref{tb:case:rel} present the top-3 nearest neighbors for the entity ``Utah'' and the relation ``relative'', respectively. We observe that the retrieved neighbors exhibit strong semantic similarity to the targets. For the entity ``Utah'', the top-3 similar entities are ``Arizona'', ``New Mexico'', and ``Colorado''; notably, these are not only US states like ``Utah'' but are also its geographical neighbors. Similarly, for the relation ``relative'', the top-3 relations pertain to kinship and familial ties. These results suggest that the representations computed by \ours capture meaningful semantic properties.

\clearpage

\section{Details of the Proposed Baselines for Fact Generation}
\label{app:fgbase}
Details of the baselines for fact generation in Section~\ref{subsec:fg} are provided here. For all LLM-based baselines, entity and relation IDs were converted to natural language descriptions for prompting. We utilized GPT-5.2 and Gemini 3.0 Pro with a default temperature of $1.0$. For baselines utilizing \maypl~\cite{maypl}, we employed an instance of \maypl retrained to support both entity and relation predictions. Note that we refer to the union of training and validation sets for ``base sets" on \wdk and \wppm, while we strictly refer to the training set for \wpp.

\begin{figure}
\begin{tcolorbox}[colback=backcolor, colframe=framecolor, title=Re-ranking Prompt Template]
\textbf{\textcolor{usercolor}{User:}} \\
You are an expert for Knowledge Graph Completion tasks.
Your goal is to perform link prediction. This involves filling in a missing element (denoted as [MASK]) in a hyper relational fact.
The missing element could be an Entity or a Relation.
Given a goal statement with a [MASK] and a list of candidate answers, you need to rank the candidates based on plausibility.
If you understand your responsibility, respond ``Yes". Otherwise, respond ``No". Do not output anything except ``Yes" and ``No".

\vspace{2mm} 
\textbf{\textcolor{modelcolor}{Model:}} \\
Yes

\tcbline

\textbf{\textcolor{usercolor}{User:}} \\
To sort the candidate answers, you need to refer to other examples that may be similar or related to it.
Some of the given examples are similar to the goal statement. You should draw analogies from them to understand the potential meaning of the goal statement.
Other provided facts contain supplementary information; capture this extra information and mine potential relationships among them to help the sorting.
Please carefully read, analyze, and reflect on these examples. Identify the reasoning patterns demonstrated in these examples and retain any information that may help your verification task.
While I provide examples, please remain silent until I ask you to respond.

\vspace{2mm} 
\textbf{\textcolor{modelcolor}{Model:}} \\
I have received your instructions and the goal statement.

\tcbline

\textbf{\textcolor{usercolor}{User:}} \\
Examples used for analogy: \\
Predict the [MASK] from the given ``\textcolor{examplecolor}{\texttt{\{{analogy\_fact\#1\}}}}''. The answer is \textcolor{examplecolor}{\texttt{\{{ans\}}}}, so the [MASK] is \textcolor{examplecolor}{\texttt{\{{ans\}}}}.\\
$\cdots$\\

Examples used to supplement information: \\
Predict the [MASK] from the given ``\textcolor{examplecolor}{\texttt{\{{supplement\_fact\#1\}}}}''. The answer is \textcolor{examplecolor}{\texttt{\{{ans\}}}}, so the [MASK] is \textcolor{examplecolor}{\texttt{\{{ans\}}}}.\\
$\cdots$\\

Keep thinking, but DO NOT give me any feedback.

\vspace{2mm} 
\textbf{\textcolor{modelcolor}{Model:}} \\
I have analyzed the examples provided and will keep thinking without giving feedback.

\tcbline

\textbf{\textcolor{usercolor}{User:}} \\
    The list of candidate answers is ``\textcolor{examplecolor}{\texttt{\{{candidate\_answer\}}}}'' and the question is predict the [MASK] from the given ``\textcolor{examplecolor}{\texttt{\{{masked\_fact\}}}}''.
    The goal is to verify the fact ``\textcolor{examplecolor}{\texttt{\{{masked\_fact\}}}}''. Based on the previous examples and your own knowledge, determine the single most probable answer from the candidate list.
    Output ONLY the index of the best candidate based on the original list order.

\vspace{2mm} 
\textbf{\textcolor{modelcolor}{Model:}} \\
0/1/2/3/4

\end{tcolorbox}
\captionof{figure}{The prompt used for fact generation across all settings in the \textit{Re-ranking} baseline.}
\label{fig:rr1}
\end{figure}

\paragraph{Iterative Prediction} This baseline represents a straightforward adaptation of a discriminative HKG model for generation. All missing components are randomly initialized and sequentially replaced with the top-1 prediction of \maypl, obtained by masking the target component. We perform 11 cycles of updates following~\cite{gibbs}, where the first 10 cycles serve as a burn-in period, and the state after the final cycle is taken as the output.

\paragraph{Gibbs Sampling} Adapted from the text generation method in~\cite{gibbs}, this baseline employs Gibbs sampling. Starting from a fact where missing components are randomly initialized while known components are preserved, we perform 11 cycles of updates. The first 10 cycles serve as burn-in, where each missing component is sequentially masked and replaced by probability-based random sampling from the top-5 predictions of \maypl. The state after the final cycle was taken as the output.

\paragraph{Re-ranking} This baseline adapts the retrieve-then-rerank strategy utilized by LLMs for KG link prediction. Specifically, it first utilizes a fact from the base set that satisfies the query to initialize the missing components. If no such fact exists, the missing components are randomly initialized. Then, for each originally missing component, \maypl suggests the top-5 candidates, from which the LLM selects the best fit to construct the final fact. The prompts are adapted from~\cite{kicgpt}; Figure~\ref{fig:rr1} illustrates the prompt used across all settings.

\paragraph{Neighbor Sets} Inspired by the neighbor-based augmentation strategy of \mukdc~\cite{mukdc}, this baseline utilizes all entities and relations sharing a fact with any known component within the query for generation. In the \textit{Scratch} setting where no known components exist, a random entity is sampled to initiate the process. We provide the prompt templates for each setting in Figures~\ref{fig:nb1},~\ref{fig:nb2}, and~\ref{fig:nb3}.

\begin{figure}
\begin{tcolorbox}[colback=backcolor, colframe=framecolor, title=Neighbor Sets Prompt Template - Scratch]
Role: You are an Expert Hyper-relational Fact Generator.\\

Task: Generate a hyper-relational fact with length \textcolor{examplecolor}{\texttt{\{{fact\_length}\}}} containing the \textcolor{examplecolor}{\texttt{\{{entities\_list\}}}} and \textcolor{examplecolor}{\texttt{\{{relations\_list\}}}}\\

Constraint: Each fact must contain exactly \textcolor{examplecolor}{\texttt{\{{fact\_length}\}}} elements. Only use the given entities and relations.\\

Output format: Subject ; Relation ; Object ; Qualifier relation ; Qualifier entity ; ...

\end{tcolorbox}
\captionof{figure}{The prompt used for the \textit{Neighbor Sets} baseline in the \textit{Scratch} setting.}
\label{fig:nb1}
\end{figure}

\begin{figure}
\begin{tcolorbox}[colback=backcolor, colframe=framecolor, title=Neighbor Sets Prompt Template - Targeted]
Role: You are an Expert Hyper-relational Fact Generator.\\

Available entities: \textcolor{examplecolor}{\texttt{\{{entities\_list\}}}}\\
Available relations: \textcolor{examplecolor}{\texttt{\{{relations\_list\}}}}\\

Task: Generate \textcolor{examplecolor}{\texttt{\{{num\_candidates\}}}} DIFFERENT hyper-relational facts by completing the following MASKED fact. Each fact should be unique.\\

Masked Fact: \textcolor{examplecolor}{\texttt{\{{masked\_fact\}}}}\\

Constraint: \\
- Each fact must contain exactly \textcolor{examplecolor}{\texttt{\{{fact\_length\}}}} elements and have the same number of elements as the masked fact\\
- Position \textcolor{examplecolor}{\texttt{\{{anchor\_idx\}}}} MUST be \textcolor{examplecolor}{\texttt{\{{anchor\_element\}}}} (this is fixed and cannot be changed)\\
- Only use entities and relations that appear in the context facts above\\
- Generate \textcolor{examplecolor}{\texttt{\{{num\_candidates\}}}} DIFFERENT facts\\

Output format: Provide exactly \textcolor{examplecolor}{\texttt{\{{num\_candidates\}}}} facts, one per line:\\
Subject ; Relation ; Object ; Qualifier relation ; Qualifier entity ; ...\\
Subject ; Relation ; Object ; Qualifier relation ; Qualifier entity ; ...\\
...\\

Provide ONLY \textcolor{examplecolor}{\texttt{\{{num\_candidates\}}}} completed facts, no explanations.

\end{tcolorbox}
\captionof{figure}{The prompt used for the \textit{Neighbor Sets} baseline in the \textit{Targeted} setting.}
\label{fig:nb2}
\end{figure}

\begin{figure}
\begin{tcolorbox}[colback=backcolor, colframe=framecolor, title=Neighbor Sets Prompt Template - Arbitrary Masking]
Role: You are an Expert Hyper-relational Fact Generator.\\

Available entities: \textcolor{examplecolor}{\texttt{\{{entities\_list\}}}}\\
Available relations: \textcolor{examplecolor}{\texttt{\{{relations\_list\}}}}\\

Task: Generate \textcolor{examplecolor}{\texttt{\{{num\_candidates\}}}} DIFFERENT hyper-relational facts by completing the following MASKED fact. Each fact should be unique.\\

Masked Fact: \textcolor{examplecolor}{\texttt{\{{masked\_fact\}}}}\\

Constraints: \\
- Each fact must contain exactly \textcolor{examplecolor}{\texttt{\{{fact\_length\}}}} elements and have the same number of elements as the masked fact\\
- The following positions are FIXED and CANNOT be changed:\\
\textcolor{examplecolor}{\texttt{\{{anchor\_information\}}}}\\
- Only use entities and relations that appear in the available entities and relations above\\
- Generate \textcolor{examplecolor}{\texttt{\{{num\_candidates\}}}} DIFFERENT facts\\

Output format: Provide exactly \textcolor{examplecolor}{\texttt{\{{num\_candidates\}}}} facts, one per line:\\
Subject ; Relation ; Object ; Qualifier relation ; Qualifier entity ; ...\\
Subject ; Relation ; Object ; Qualifier relation ; Qualifier entity ; ...\\
...\\

Provide ONLY \textcolor{examplecolor}{\texttt{\{{num\_candidates\}}}} completed facts, no explanations.

\end{tcolorbox}
\captionof{figure}{The prompt used for the \textit{Neighbor Sets} baseline in the \textit{Arbitrary Masking} setting}
\label{fig:nb3}
\end{figure}

\paragraph{Few-shot Facts} This baseline is similar to the \textit{Neighbor Sets} baseline, but instead utilizes 30 training facts involving the known components as few-shot examples. Figures~\ref{fig:ego1},~\ref{fig:ego2}, and~\ref{fig:ego3} present the prompts utilized for the \textit{Scratch}, \textit{Targeted}, and \textit{Arbitrary Masking} settings, respectively.

\begin{figure}
\begin{tcolorbox}[colback=backcolor, colframe=framecolor, title=Few-shot Facts Prompt Template - Scratch]
Role: You are an Expert Hyper-relational Fact Generator.\\

Here are \textcolor{examplecolor}{\texttt{\{{num\_facts\}}}} existing facts for context:\\
\textcolor{examplecolor}{\texttt{\{{facts\}}}}\\

Task: Generate a NEW hyper-relational fact with length \textcolor{examplecolor}{\texttt{\{{fact\_length\}}}}\\

Constraint: Each fact must contain exactly \textcolor{examplecolor}{\texttt{\{{fact\_length\}}}} elements. Only use use entities and relations that appear in the facts above.\\

Output format: Subject ; Relation ; Object ; Qualifier relation ; Qualifier entity ; ...

\end{tcolorbox}
\captionof{figure}{The prompt used for the \textit{Few-shot Facts} baseline in the \textit{Scratch} setting.}
\label{fig:ego1}
\end{figure}

\begin{figure}
\begin{tcolorbox}[colback=backcolor, colframe=framecolor, title=Few-shot Facts Prompt Template - Targeted]
Role: You are an Expert Hyper-relational Fact Generator.\\

Here are \textcolor{examplecolor}{\texttt{\{{num\_facts\}}}} existing facts for context:\\
\textcolor{examplecolor}{\texttt{\{{facts\}}}}\\

Task: Generate \textcolor{examplecolor}{\texttt{\{{num\_candidates\}}}} DIFFERENT hyper-relational facts by completing the following MASKED fact. Each fact should be unique.\\

Masked Fact: \textcolor{examplecolor}{\texttt{\{{masked\_facts\}}}}\\

Constraint: \\
- Each fact must contain exactly \textcolor{examplecolor}{\texttt{\{{fact\_length\}}}} elements and have the same number of elements as the masked fact\\
- Position \textcolor{examplecolor}{\texttt{\{{anchor\_index\}}}} MUST be \textcolor{examplecolor}{\texttt{\{{anchor\_element\}}}} (this is fixed and cannot be changed)\\
- Only use entities and relations that appear in the context facts above\\
- Generate \textcolor{examplecolor}{\texttt{\{{num\_candidates\}}}} DIFFERENT facts\\

Output format: Provide exactly \textcolor{examplecolor}{\texttt{\{{num\_candidates\}}}} facts, one per line:\\
Subject ; Relation ; Object ; Qualifier relation ; Qualifier entity ; ...\\
Subject ; Relation ; Object ; Qualifier relation ; Qualifier entity ; ...\\
...\\

Provide ONLY \textcolor{examplecolor}{\texttt{\{{num\_candidates\}}}} completed facts, no explanations.

\end{tcolorbox}
\captionof{figure}{The prompt used for the \textit{Few-shot Facts} baseline in the \textit{Targeted} setting.}
\label{fig:ego2}
\end{figure}

\begin{figure}
\begin{tcolorbox}[colback=backcolor, colframe=framecolor, title=Few-shot Facts Prompt Template - Arbitrary Masking]
Role: You are an Expert Hyper-relational Fact Generator.\\

Here are \textcolor{examplecolor}{\texttt{\{{num\_facts\}}}} existing facts for context:\\
\textcolor{examplecolor}{\texttt{\{{facts\}}}}\\

Task: Generate \textcolor{examplecolor}{\texttt{\{{num\_candidates\}}}} DIFFERENT hyper-relational facts by completing the following MASKED fact. Each fact should be unique.\\

Masked Fact: \textcolor{examplecolor}{\texttt{\{{masked\_facts\}}}}\\

Constraints: \\
- Each fact must contain exactly \textcolor{examplecolor}{\texttt{\{{fact\_length\}}}} elements and have the same number of elements as the masked fact\\
- The following positions are FIXED and CANNOT be changed:\\
\textcolor{examplecolor}{\texttt{\{{anchor\_information\}}}}\\
- Only use entities and relations that appear in the context facts above\\
- Generate \textcolor{examplecolor}{\texttt{\{{num\_candidates\}}}} DIFFERENT facts\\

Output format: Provide exactly \textcolor{examplecolor}{\texttt{\{{num\_candidates\}}}} facts, one per line:\\
Subject ; Relation ; Object ; Qualifier relation ; Qualifier entity ; ...\\
Subject ; Relation ; Object ; Qualifier relation ; Qualifier entity ; ...\\
...\\

Provide ONLY \textcolor{examplecolor}{\texttt{\{{num\_candidates\}}}} completed facts, no explanations.

\end{tcolorbox}
\captionof{figure}{The prompt used for the \textit{Few-shot Facts} baseline in the \textit{Arbitrary Masking} setting.}
\label{fig:ego3}
\end{figure}

\paragraph{Random Facts} The LLM is provided with a global pool of 1,000 facts randomly sampled from the base set and instructed to generate new facts using only the observed entities and relations. The prompt for the \textit{Scratch} setting is shown in Figure~\ref{fig:fs1}, whereas Figure~\ref{fig:fs2} and Figure~\ref{fig:fs3} shows the prompt for the \textit{Targeted} and \textit{Arbitrary Masking} settings, respectively.

\begin{figure}
\begin{tcolorbox}[colback=backcolor, colframe=framecolor, title=Random Facts Prompt Template - Scratch]
Role: You are an Expert Hyper-relational Fact Generator.\\
        
Here are \textcolor{examplecolor}{\texttt{\{{num\_facts\}}}} hyper-relational facts for context:\\
\textcolor{examplecolor}{\texttt{\{{facts\}}}}\\

Task: Generate \textcolor{examplecolor}{\texttt{\{{generate\_size\}}}} NEW facts with target length \textcolor{examplecolor}{\texttt{\{{fact\_length\}}}} based on this list.\\
Use entities, relations, and qualifier key-values only from the provided list.\\

Constraint: Each fact must contain exactly \textcolor{examplecolor}{\texttt{\{{fact\_length\}}}} elements.\\

Output format: Subject ; Relation ; Object ; Qualifier relation ; Qualifier entity ; ...

\end{tcolorbox}
\captionof{figure}{The prompt used for the \textit{Random Facts} baseline in the \textit{Scratch} setting.}
\label{fig:fs1}
\end{figure}

\begin{figure}
\begin{tcolorbox}[colback=backcolor, colframe=framecolor, title=Random Facts Prompt Template - Targeted]
Role: You are an Expert Hyper-relational Fact Generator.\\
Here are \textcolor{examplecolor}{\texttt{\{{num\_facts\}}}} hyper-relational facts for context:\\
\textcolor{examplecolor}{\texttt{\{{facts\}}}}\\

Task: Complete the following \textcolor{examplecolor}{\texttt{\{{generate\_size\}}}} MASKED facts. For EACH masked fact, generate \textcolor{examplecolor}{\texttt{\{{num\_candidates\}}}} DIFFERENT hyper-relational facts by completing the following MASKED fact. Each fact should be unique.\\

Masked facts to complete:\\
\textcolor{examplecolor}{\texttt{\{{masked\_fact\}}}}\\

Constraint:\\
- For each masked fact, maintain its exact length\\
- Keep non-MASK elements in their original positions\\
- Only use entities and relations from the context facts above\\
- Generate \textcolor{examplecolor}{\texttt{\{{num\_candidates\}}}} different completions for EACH masked fact\\

Output format: For each masked fact, provide \textcolor{examplecolor}{\texttt{\{{num\_candidates\}}}} completions, one per line:\\
Fact 1 - Completion 1: Subject ; Relation ; Object ; ...\\
Fact 1 - Completion 2: Subject ; Relation ; Object ; ...\\
...\\
Fact 1 - Completion \textcolor{examplecolor}{\texttt{\{{num\_candidates\}}}}: Subject ; Relation ; Object ; ...\\
Fact 2 - Completion 1: Subject ; Relation ; Object ; ...\\
...\\

Provide ONLY completed facts in the format above, no explanations.

\end{tcolorbox}
\captionof{figure}{The prompt used for the \textit{Random Facts} baseline in the \textit{Targeted} settings.}
\label{fig:fs2}
\end{figure}

\begin{figure}
\begin{tcolorbox}[colback=backcolor, colframe=framecolor, title=Random Facts Prompt Template - Arbitrary Masking]
Role: You are an Expert Hyper-relational Fact Generator.\\
Here are \textcolor{examplecolor}{\texttt{\{{num\_facts\}}}} hyper-relational facts for context:\\
\textcolor{examplecolor}{\texttt{\{{facts\}}}}\\

Your task is to complete the following \textcolor{examplecolor}{\texttt{\{{generate\_size\}}}} MASKED facts. For EACH masked fact, generate \textcolor{examplecolor}{\texttt{\{{num\_candidates\}}}} DIFFERENT hyper-relational facts. \\Each fact should be unique.\\

Masked facts to complete:\\
\textcolor{examplecolor}{\texttt{\{{masked\_fact\}}}}\\

Constraints:\\
- For each masked fact, maintain its exact length\\
- Keep non-MASK elements in their original positions\\
- Only use entities and relations from the context facts above\\
- Generate \textcolor{examplecolor}{\texttt{\{{num\_candidates\}}}} different completions for EACH masked fact\\

Output format: For each masked fact, provide \textcolor{examplecolor}{\texttt{\{{num\_candidates\}}}} completions, one per line:\\
Fact 1 - Completion 1: Subject ; Relation ; Object ; ...\\
Fact 1 - Completion 2: Subject ; Relation ; Object ; ...\\
...\\
Fact 1 - Completion \textcolor{examplecolor}{\texttt{\{{num\_candidates\}}}}: Subject ; Relation ; Object ; ...\\
Fact 2 - Completion 1: Subject ; Relation ; Object ; ...\\
...\\

Provide ONLY completed facts in the format above, no explanations.

\end{tcolorbox}
\captionof{figure}{The prompt used for the \textit{Random Facts} baseline in the \textit{Arbitrary Masking} settings.}
\label{fig:fs3}
\end{figure}

\paragraph{Autoregressive} Generation proceeds autoregressively starting from the head entity. If no known components are given, a random entity is used as the starting head. The provided examples are dynamic: the first step uses 30 examples containing known components, while subsequent steps retrieve 30 examples involving the immediately preceding predicted element. The specific prompts for the \textit{Scratch}, \textit{Targeted}, and \textit{Arbitrary Masking} settings are illustrated in Figures~\ref{fig:seq1},~\ref{fig:seq2}, and~\ref{fig:seq3}, respectively.

\begin{figure}
\begin{tcolorbox}[colback=backcolor, colframe=framecolor, title=Autoregressive Prompt Template - Scratch]
\textbf{\textcolor{usercolor}{User:}} \\
Role: You are an Expert Hyper-relational Fact Generator.\\

The HEAD entity has been selected: \textcolor{examplecolor}{\texttt{\{{head\}}}}

Here are facts where ``\textcolor{examplecolor}{\texttt{\{{head\}}}}" appears:\\
\textcolor{examplecolor}{\texttt{\{{head\_examples\}}}}\\

Task: Select ONE relation from the facts above that could connect to this head entity.\\
You must only use relations that appear in the provided facts.

Output ONLY the relation name, nothing else.

\vspace{2mm} 
\textbf{\textcolor{modelcolor}{Model:}} \\
\textcolor{examplecolor}{\texttt{\{{relation\}}}}

\tcbline

\textbf{\textcolor{usercolor}{User:}} \\
Current fact being built: \textcolor{examplecolor}{\texttt{\{{current\_fact\}}}}\\
Here are facts where ``\textcolor{examplecolor}{\texttt{\{{relation\}}}}" appears:\\
\textcolor{examplecolor}{\texttt{\{{relation\_examples\}}}}\\

Task: Select ONE entity from the facts above that could connect to this relation.\\
You must only use entities that appear in the provided facts and the facts shown in previous steps.

Output ONLY the entity name, nothing else.

\vspace{2mm} 
\textbf{\textcolor{modelcolor}{Model:}} \\
\textcolor{examplecolor}{\texttt{\{{tail\}}}}

\tcbline

\textbf{\textcolor{usercolor}{User:}} \\
Current fact being built: \textcolor{examplecolor}{\texttt{\{{current\_fact\}}}}\\
Here are facts where ``\textcolor{examplecolor}{\texttt{\{{last\_element\}}}}" appears:\\
\textcolor{examplecolor}{\texttt{\{{last\_element\_examples\}}}}\\

Task: Select ONE relation from the facts above that could connect to this entity.\\
You must only use relations that appear in the provided facts and the facts shown in previous steps.

Output ONLY the relation name, nothing else.

\vspace{2mm} 
\textbf{\textcolor{modelcolor}{Model:}} \\
\textcolor{examplecolor}{\texttt{\{{qualifier\_relation\}}}}

\tcbline

\textbf{\textcolor{usercolor}{User:}} \\
Current fact being built: \textcolor{examplecolor}{\texttt{\{{current\_fact\}}}}\\
Here are facts where ``\textcolor{examplecolor}{\texttt{\{{qualifier\_relation\}}}}" appears:\\
\textcolor{examplecolor}{\texttt{\{{qualifier\_relation\_examples\}}}}\\

Task: Select ONE entity from the facts above that could connect to this relation.\\
You must only use entities that appear in the provided facts and the facts shown in previous steps.

Output ONLY the entity name, nothing else.

\vspace{2mm} 
\textbf{\textcolor{modelcolor}{Model:}} \\
\textcolor{examplecolor}{\texttt{\{{qualifier\_entity\}}}}
\tcbline
$\cdots$

\end{tcolorbox}
\captionof{figure}{The prompt used for the \textit{Autoregressive} baseline in the \textit{Scratch} setting.}
\label{fig:seq1}
\end{figure}

\begin{figure}
\begin{tcolorbox}[colback=backcolor, colframe=framecolor, title=Autoregressive Prompt Template - Targeted]
\textbf{\textcolor{usercolor}{User:}} \\
Role: You are an Expert Hyper-relational Fact Generator.\\

Current fact being built: \textcolor{examplecolor}{\texttt{\{{current\_fact\}}}}\\

Here are facts where ``\textcolor{examplecolor}{\texttt{\{{last\_element\}}}}" appears:\\
\textcolor{examplecolor}{\texttt{\{{last\_element\_example\}}}}\\

Task: Select ONE \textcolor{examplecolor}{\texttt{\{{element\_type\}}}} from the facts above that could fill the [FILLING NOW] position.\\
You must only use \textcolor{examplecolor}{\texttt{\{{element\_type\}}}} that appear in the provided facts and the facts shown in previous steps.\\

Do not create same facts as in the examples.\\

Output ONLY the \textcolor{examplecolor}{\texttt{\{{element\_type\}}}} name, nothing else.

\vspace{2mm} 
\textbf{\textcolor{modelcolor}{Model:}} \\
\textcolor{examplecolor}{\texttt{\{{element\}}}}
\tcbline
$\cdots$
\end{tcolorbox}
\captionof{figure}{The prompt used for the \textit{Autoregressive} baseline in the \textit{Targeted} setting.}
\label{fig:seq2}
\end{figure}

\begin{figure}
\begin{tcolorbox}[colback=backcolor, colframe=framecolor, title=Autoregressive Prompt Template - Arbitrary Masking]
\textbf{\textcolor{usercolor}{User:}} \\
Role: You are an Expert Hyper-relational Fact Generator.\\

Current fact being built: \textcolor{examplecolor}{\texttt{\{{current\_fact\}}}}\\
Fixed positions (CANNOT change): \textcolor{examplecolor}{\texttt{\{{anchor\_information\}}}}\\

Here are facts for context:\\
\textcolor{examplecolor}{\texttt{\{{context\_fact\}}}}\\

Task: Select ONE \textcolor{examplecolor}{\texttt{\{{element\_type\}}}} from the facts above that could fill the [FILLING NOW] position.\\
You must only use \textcolor{examplecolor}{\texttt{\{{element\_type\}}}} that appear in the provided facts and the facts shown in previous steps.\\

Do not create same facts as in the examples.\\

Output ONLY the \textcolor{examplecolor}{\texttt{\{{element\_type\}}}} name, nothing else.

\vspace{2mm} 
\textbf{\textcolor{modelcolor}{Model:}} \\
\textcolor{examplecolor}{\texttt{\{{element\}}}}
\tcbline
$\cdots$
\end{tcolorbox}
\captionof{figure}{The prompt used for the \textit{Autoregressive} baseline in the \textit{Arbitrary Masking} setting.}
\label{fig:seq3}
\end{figure}

We provide the implementation of the baseline methods for fact generation at \url{https://github.com/bdi-lab/KREPE}.


\end{document}